%% file: emnlp2019.tex
\documentclass[11pt,a4paper]{article}
\usepackage[hyperref]{emnlp-ijcnlp-2019}
\usepackage{times}
\usepackage{latexsym}
\usepackage{verbatim}

\usepackage{amsmath}
\usepackage{amsfonts}

\usepackage{graphicx}
\usepackage{multirow}
\usepackage{booktabs}

\usepackage{url}

\aclfinalcopy % Uncomment this line for the final submission
\title{CraftAssist Instruction Parsing: Semantic Parsing for a Minecraft Assistant}

\author{Yacine Jernite* \qquad 
  Kavya Srinet* 
   \qquad
   Jonathan Gray 
    \qquad
      Arthur Szlam \\  \\
      Facebook AI Research
   }

\date{}

\begin{document}
\maketitle
\begin{abstract}
We propose a large scale semantic parsing dataset focused on instruction-driven communication with an agent in Minecraft. We describe the data collection process which yields additional 35K human generated instructions with their semantic annotations. We report the performance of three baseline models and find that while a dataset of this size helps us train a usable instruction parser, it still poses interesting generalization challenges which we hope will help develop better and more robust models.
\end{abstract}

\input{introduction}

\input{related}

\input{grammar}
\input{data}

\input{modeling}
\input{experiments}

\section{Conclusion}
In this work, we have described a grammar over a control system for a Minecraft assistant. We then discussed the creation of a dataset of natural language utterances with associated logical forms from this grammar that can be executed in-game. Finally, we showed the results of using this new dataset to train several neural models for parsing natural language instructions. We find that the models we trained were able to fit the templated data nearly perfectly and the rephrased data with some accuracy, but struggled to adapt to the human-generated data. In our view, the problem of using the small number of annotated (grammar-free) human data with the infinite generations of our grammar to improve results on human-distribution to be an exciting area of research. %and hope that future work tackling the task presented here may produce better results.

\newpage

% include your own bib file like this:
%\bibliographystyle{acl}
%\bibliography{acl2018}
\bibliography{refs}
\bibliographystyle{acl_natbib}

\clearpage

\appendix

\label{sec:supplemental}

\input{action_tree}

\input{task_instructions}

\input{confusion}

\end{document}

%% file: introduction.tex
\section{Introduction} 
Semantic parsing is used as a component for natural language understanding in human-robot interaction systems \cite{tellex2011understanding, matuszek2013learning}, and for virtual assistants  \cite{kollar2018alexa}. Recently, researchers have shown success with deep learning methods for semantic parsing, e.g. \cite{dong2016language, jia2016data, zhong2017seq2sql}. However, to fully utilize powerful neural network approaches, it is necessary to have large numbers of training examples. In the space of human-robot (or human-assistant) interaction, the publicly available semantic parsing datasets are small. Furthermore, it can be difficult to reproduce the end-to-end results (from utterance to action) because of the wide variety of robot setups and proprietary nature of personal assistants.

In this work, we introduce a new semantic parsing dataset for human-bot interactions. Our ``robot'' or ``assistant''  is embodied in the sandbox construction game Minecraft\footnote{\url{https://minecraft.net/en-us/}.  We limit ourselves to creative mode for this work}, a popular multiplayer open-world voxel-based crafting game. We also provide the associated platform for executing the logical forms in game.

Situating the assistant in Minecraft has several benefits for studying task oriented natural language understanding (NLU). Compared to physical robots, Minecraft allows less technical overhead irrelevant to NLU, such as difficulties with hardware and large scale data collection. On the other hand, our bot has all the basic in-game capabilities of a player, including movement and placing or removing voxels. Thus Minecraft preserves many of the NLU elements of physical robots, such as discussions of navigation and spatial object reference. 
 
%\begin{figure*}[t!]
%    \centering
%%     \includegraphics[natwidth=\linewidth]{figures/ActionGrammar.png}
%    \includegraphics[width=\linewidth]{figures/ActionGrammar.png}
%    \caption{Action space grammar}
%    \label{fig:action_grammar}
%\end{figure*}  

Furthermore, working in Minecraft may enable large scale human interaction because of its large player base, in the tens of millions. 
 %Compared with personal assistants, Minecraft offers a low
% for allows the possibility of interacting with large number of players
Although Minecraft's simulation of physics is simplified, the task space is complex.  There are many atomic objects in Minecraft, such as animals and block-types, that require no perceptual modeling. For researchers interested in the interactions between perception and language, collections of voxels making up a ``house'' or a ``hill'' are not atomic objects and the assistant cannot apprehend them without a perceptual system. 

Our contributions in the paper are as follows:
\newline
\noindent{\bf Grammar:} We develop a set of action primitives and grammar over these primitives that comprise a mid-level interface to Minecraft, for machine learning agents.  See Section \ref{sec:grammar}.
\newline 
\noindent{\bf Data:} Using a collection of language templates to convert logical forms over the primitives into pseudo-natural language, we build a dataset of language instructions with logical form annotation by having crowd-sourced workers rephrase the language outputs, as in \cite{wang2015building}.  We also collect a test set of crowd-sourced annotations of commands generated independent of our grammar. In addition to the natural language commands and the associated logical forms, we also make available the code to execute these in the game, allowing the reproduction of end-to-end results. See Section \ref{sec:data}.
%We hope in addition to use as a semantic parsing dataset, these will be useful for researchers in reinforcement learning.
\newline   
\noindent{\bf Models:} We show the results of several neural semantic parsing models trained on our data. See Section \ref{sec:modeling} and \ref{sec:experiments}.
We also will provide access to an interactive bot using these models for parsing\footnote{Instructions will be available at \url{http://craftassist.io/acl2019demo}}. %We demonstrate  that adapting the neural architecture to the grammar can improve parsing accuracy.

\begin{figure*}[t!]
    \centering
    \includegraphics[width=\linewidth ]{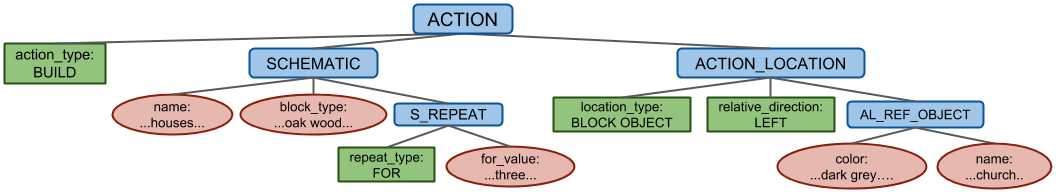}
    \caption{Parse tree for ``Make three oak wood houses to the left of the dark grey church.''    \label{fig:example_parse}
}
\end{figure*}

%Advantages of working with MC / game environment -- General usefulness of a large semantic parsing dataset in the context of a semi-tractable virtual environment.

%% file: related.tex
\section{Related Work}

%Semantic parsing -- 1970s AI -- Question answering approaches -- FSP -- Grounded Language -- Relevant MC recent work -- Alexa Meaning Representation / Challenge

There have been a number of datasets of natural language paired with logical forms to evaluate semantic parsing approaches, e.g. \cite{price1990evaluation, tang2001using, cai2013large, wang2015building, zhong2017seq2sql}. The dataset presented in this work is an order of magnitude larger than those in \cite{price1990evaluation, tang2001using, cai2013large} and is similar in scale to \cite{wang2015building, zhong2017seq2sql}. We use the data collection strategy in \cite{wang2015building} to build the pairings between logical forms and natural language: first building the grammar, then generating from the grammar via templates, and then using crowd-sourced workers to rephrase the templated generations. However, we also collect a test set of ``free'' commands and use crowd-sourced workers to annotate these.

In addition to connecting natural language to logical forms, our dataset connects both of these to a dynamic environment.  
In \cite{tellex2011understanding, matuszek2013learning} semantic parsing has been used for interpreting natural language commands for robots. %, matuszek2013learning}.  
In our paper, the ``robot'' is embodied in the Minecraft game instead of in the physical world.
%\cite{matuszek2013learning} :~500 examples, RCL differences (but similar- we have loops etc.), deep models instead of feature reps, templates

Semantic parsing in a voxel-world recalls \cite{wang2017naturalizing}, where the authors describe a method for building up a programming language from a small core via interactions with players.  %In this work, we made a substantial effort to build coverage ourselves via a corpus of templates, but 

%\cite{wang2016learning, wang2017naturalizing} %maybe second cite should go to discussion

We demonstrate the results of several neural parsing models on our dataset.  In particular, we show the results of a reimplementation of \cite{dong2016language} adapted to our grammar. There have been several other papers proposing neural architectures for semantic parsing, for example \cite{jia2016data, zhong2017seq2sql}.  In those papers, as in this one, the models are trained with full supervision of the mapping from natural language to logical forms, without considering the results of executing the logical form (in this case, the effect on the environment of executing the actions denoted by the logical form). There has been progress towards ``weakly supervised'' semantic parsing
% (with or without deep architectures ) 
 \cite{artzi2013weakly, liang2016neural, guu2017language} where the logical forms are hidden variables, and the only supervision given is the result of executing the logical form. There are now approaches that have shown promise without even passing through (discrete) logical forms at all \cite{riedel2016programming, neelakantan2016learning}.  We hope that the dataset introduced here, which has supervision at the level of the logical forms, but whose underlying grammar and environment can be used to generate essentially infinite weakly supervised or execution rewards, will also be useful for studying these models.

Minecraft, especially via the MALMO project \cite{johnson2016malmo} has been used as a base environment for several machine learning papers.  
Often Minecraft is used as a testbed for reinforcment learning \cite{shu2017hierarchical,udagawa2016fighting,alaniz2018deep,oh2016control,tessler2017deep}.
In these papers, the agent is trained to complete tasks by issuing low level actions (as opposed to our higher level primitives) and receiving a reward
on success.  Some of these papers(e.g. \cite{oh2017zero}) do consider simplified, templated language as a method for composably specifying tasks, but
training an RL agent to execute the scripted primitives in our grammar is already nontrivial, and so the task space and language is more constrained 
than what we use here.  Nevertheless, our work may be useful to researchers interested in RL- using our grammar and executing in game can supply (hard) tasks and descriptions.
Another set of papers 
\cite{kitaev2017misty, yi2018neural}  have used Minecraft for visual question answering with logical forms. Our work extends these to interactions with the environment.
Finally, \cite{allison2018players} is a more focused study on how a human might interact with a Minecraft agent; our collection of free generations (see \ref{sec:prompts}) includes annotated 
examples from similar studies of players interacting with a player pretending to be a bot.

%% file: grammar.tex
\section{A Natural Language Interface}
\label{sec:grammar}

We want to interpret natural language commands given to an agent with a pre-defined set of capabilities. We start by providing an overview of these capabilities and the action space that they entail, then define a grammar to capture this action space.

\subsection{Agent Action Space}

The goal of the proposed agent is to help a player create structures and mechanisms in a voxelized world by moving around, and placing and removing blocks. To this end, the agent needs to be able to understand a number of high-level commands, which we present here.

\paragraph{Basic action commands} First, we need commands corresponding to high level actions of the agent. For example, we may ask it to \textsc{build} an object from a known schematic or to copy an existing structure at a given location, or to \textsc{destroy} one. Similarly, it might be useful to be able to ask the agent to \textsc{dig} a hole of a given shape at a specified location, or on the contrary to \textsc{fill} one up. The agent can also be asked to complete an already structure however it sees fit (this action is called \textsc{freebuild}), or to \textsc{spawn} game mobs. Finally, we need to be able to direct the agent to \textsc{move} to a location.

\paragraph{Teaching and querying the bot} In order to understand most of the above commands, the agent needs to have an internal representation of the world. We want to be able to add to this representation by allowing the user to \textsc{tag} existing objects with names or properties.  This can be considered a basic version of the self-improvement capabilities in \cite{kollar2013toward, thomason2015learning, wang2016learning, wang2017naturalizing}. Conversely, to query this internal state, we can ask the agent to \textsc{answer} questions about the world.  This part of the grammar is similar to the visual question-answering in \cite{yi2018neural}

\paragraph{Control commands} Additionally, we want to be able to ask the agent to \textsc{stop} or \textsc{resume} an action, or to \textsc{undo} the result of a recent command.  Finally, the agent needs to be able to understand when a sentence does not correspond to any of the above mentioned actions, and map it to a \textsc{noop} command.

\subsection{Parsing Grammar}

All of the above commands are represented as trees encoding all of the information necessary for their execution. Figure~\ref{fig:example_parse} presents an example parse tree for the \textsc{build} command \textit{``Make three oak wood houses to the left of the dark grey church.''}

\paragraph{Internal nodes} Each action has a set of possible arguments, which themselves have a recursive argument structure. Each of these action types and complex arguments corresponds to an internal node (blue rounded rectangles in Figure~\ref{fig:example_parse}), with its children providing more specific information. For example, the \textsc{build} action can specify a \textsc{schematic} (what we want to build) and a \textsc{location} child (where we want to build it). In turn, the \textsc{schematic} can specify a general category (house, bridge, temple, etc\ldots), as well as a set of properties (size, color, building material, etc...), and in our case also has a \textsc{repeat} child subtree specifying how many we want to build. Similarly, the \textsc{location} can specify an absolute location, a distance, direction,  and information about the location \textsc{reference object} stored in a child subtree.

One notable feature of this representation is that we do not know \textit{a priori} which of a node's possible children will be specified. For example, \textsc{build} can have a \textsc{schematic} and a \textsc{location} specified (\textit{``Build a house over there.''}), just a \textsc{schematic} (\textit{``Build a house.''}), just a \textsc{location} (\textit{``Build something next to the bridge.''}), or neither (\textit{``Make something.''}).

The full grammar is specified in Figure~\ref{fig:action_grammar}. In addition to the various \textsc{location}, \textsc{reference object}, \textsc{schematic}, and \textsc{repeat} nodes which can be found at various levels, another notable subtree is the action's \textsc{stop condition}, which essentially allows the agent to understand ``while'' loops (for example: ``dig down until you hit the bedrock'' or ``follow me'').

\paragraph{Leaf nodes} Eventually, arguments have to be specified in terms of values which correspond to agent primitives. We call these nodes categorical leaves (green rectangles in Figure~\ref{fig:example_parse}). The root of the tree has a categorical leaf child which specifies the \textbf{action type}, \textsc{build} in our example. There are also nodes specifying the \textbf{repeat type} in the \textsc{repeat} sub-tree ("make three houses" corresponds to executing a \textsc{for} loop), the \textsc{location type} (the location is given in reference to the \textsc{block object} that is the ``dark grey church''), and the \textbf{relative direction} to the reference, here \textsc{left}.

However, there are limits to what we can represent with a pre-specified set of hard-coded primitives, especially if we want our agent to be able to learn new concepts or new values. Additionally, even when there is a pre-specified agent primitive, mapping some parts of the command to a specific value might be better left to an external module (e.g. mapping a number string to an integer value). For both of these reasons, we also have span leaves (red ovals in Figure~\ref{fig:example_parse}). This way, a model can learn to generalize to e.g. colors or size descriptions that it has never seen before. The \textsc{schematic} is specified by the command sub-string corresponding to its \textbf{name} (``houses'') and the requested \textbf{block type} (``oak wood''). The range of the for loop is specified by the \textsc{repeat}'s \textbf{for value} (``three''), and the \textsc{reference object} is denoted in the command by its generic \textbf{name} and specific \textbf{color} (``church'' and ``dark grey'').

%% file: data.tex
\section{The CAIP Dataset}
\label{sec:data}

This paper introduces the CraftAssist Instruction Parsing (CAIP) dataset of English-language commands and their associated "action trees", as defined in Section~\ref{sec:grammar} (see Appendix~\ref{sec:action_tree} for examples and a full grammar specification). CAIP is a composite dataset containing a combination of algorithmically generated commands and human-written natural language commands.

\subsection{Generated Data}
\label{sec:generated_data}

We start by algorithmically generating action trees (logical forms over the grammar) with associated surface forms through the use of templates. To that end, we first define a set of template objects, which link an atomic concept in the game world to several ways it can be described through language. For example the template object \texttt{Move} links the action type \textsc{move} to the utterances \textit{go},  \textit{walk},  \textit{move},\ldots Likewise, the template object \texttt{RelativeDirection} links all of the direction primitives to their names. Some template objects also have purely linguistic functions in order to make the sentence more natural but without referring to any information relevant to the tree. For example, the object \texttt{ALittle} can be realized into \textit{a bit},  \textit{a little}, \textit{somewhat}, \ldots 

Then, we build recursive templates for each action as recursive sequences of templates and template objects. For each of these templates, we can then sample a game value and its corresponding string. By concatenating these, we obtain an action tree and its corresponding language description. Consider for example the template [\texttt{Move}, \texttt{ALittle}, \texttt{RelativeDirection}] made up of the template objects described above. One possible realization could be the description \textit{go a little to the left} paired with an action tree specifying the action type as \textsc{move}, and an \textsc{action location} sub-tree with which a child relative direction categorical node which has value \textsc{left}. Finally, in addition to the action-specific templates, we also generate
training data for the \textsc{noop} action type by sampling dialogue lines from the Cornell Movie Dataset \cite{Danescu-Niculescu-Mizil+Lee:11a}.

We wrote 3,900 templates in total. We can create a training example for a parsing model by choosing one of them at random, and then sampling a (description, tree) pair from it, which, given the variety and modularity of the template objects, yields virtually unlimited data (for practical reasons, we pre-generate a set of 800K training, 5K validation, and 5K test examples for our experiments). The complete list of templates and template objects is included in the Supplementary Material.

% We generated a large database of natural language commands and their corresponding action trees using a system of hierarchically composable templates. Each action is characterized by a set of supported templates, each of which is an ordered list of template objects.

% To generate a command, we choose a random action from the set of supported actions (a complete list is presented in section \ref{sec:action_tree}) and then a random template is chosen from the action's template set. Each template object is then converted to text to produce the command, and key-value pairs to produce the action tree.

% An example template for the \textsc{Move} action is composed of the  objects. Given different resolutions of each object, the template might produce commands like "move a little to the right" or "can you walk a bit south". Some template objects affect the action tree (e.g. the "relative\_direction" key is different in the two examples above), while others like \texttt{ALittle} do not.

% Template objects may also make use of other template objects, making the conversion to text a recursive process. The text and action tree fragments produced by template objects that are shared across templates can be reused (e.g. both the \textsc{Move} and \textsc{Build} actions may specify a \texttt{Location}).

\begin{figure*}
\includegraphics[width=\linewidth ]{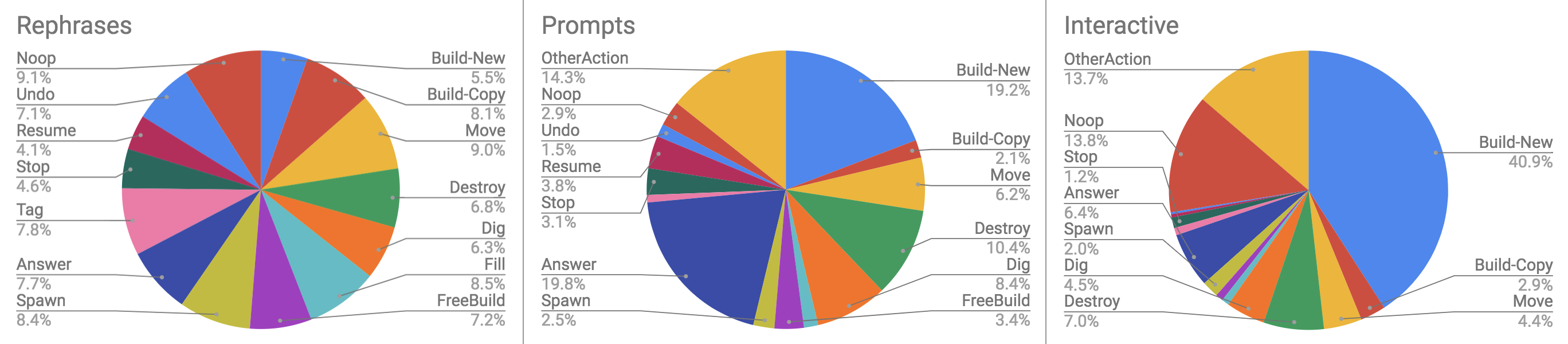}
\caption{Frequency of each action type in the different data collection schemes described in Section~\ref{sec:collected_data}. The \textsc{Build} action is divided into \textsc{Build-New} (a command to build a brand new structure, which may specify a \textsc{schematic}) and \textsc{Build-Copy} (an command to duplicate an existing structure, which specifies a \textsc{reference object}.\label{fig:action_freqs}
}
\end{figure*}

\subsection{Collected Data}
\label{sec:collected_data}

To supplement the generated data, natural language commands written by crowd-sourced workers were collected in a variety of settings.

\subsubsection{Rephrases} While the template generations yield a great variety of language, they cannot cover all possible ways of phrasing a specific instruction. In order to supplement them, we asked crowd-sourced workers to  rephrase some of the produced instructions into commands in alternate, natural English that does not change the meaning of the sentence. This setup enables the collection of unique English commands whose action trees are already known. Note that a rephrased sentence will have the same action tree structure, but the positions of the words corresponding to span nodes may change. % , and hence the values of the spans.
% For example, the command "build a pillar", represented by the action tree:  \{"Build": \{"schematic": \{"has\_name\_": [2,~2]\}\}\}, might be rephrased to "make me a pillar", where the word "pillar" has now moved to index 3. 
To account for this, words contained in a span range in the original sentence are highlighted in the task, and crowd-sourced workers are asked to highlight the corresponding words in their rephrased sentence. Then the action tree span values are substituted for the rephrased sentence to get the corresponding tree. This yields a total of 32K rephrases. We use 30K for training, 1K for validation, and 1K for testing.

% jgray: this paragraph assumes some context about there being an actual interactive bot
\subsubsection{Image and Text Prompts} \label{sec:prompts}We also presented crowd-sourced workers with a description of the capabilities of an assistant bot in a creative virtual environment (which matches the set of allowed actions in the grammar), and (optionally) some images of a bot in a game environment. They were then asked to provide examples of commands that they might issue to an in-game assistant. We refer to these instructions as ``prompts'' in the rest of this paper. The complete instructions shown to workers is included in appendix \ref{fig:freegen}.

% The process of annotating these sentences with their corresponding action trees is described in section \ref{sec:annotation_tool}.

\subsubsection{Interactive Gameplay}\label{sec:human_bot} We asked crowd-sourced workers to play creative-mode Minecraft with an assistant bot, and they were instructed to use the in-game chat to direct the bot in whatever way they chose. The exact instructions are included in appendix \ref{sec:appen}. Players in this setting had no prior knowledge of the bot's capabilities or the parsing grammar.
% Each line of in-game chat written by a player in this scenario was annotated with its corresponding action tree by the process described in section \ref{sec:annotation_tool}.

% \subsubsection{Annotation tool}
% \label{sec:annotation_tool}

\subsubsection{Annotation Tool}
Both prompts and interactive instructions come without a reference tree and need to be annotated. To facilitate this process,
% the annotation of a natural language command with its corresponding action tree representation,
 we designed a web-based tool which asks users a series of multiple-choice questions to determine the semantic content of a sentence. The responses to some questions will prompt other more specific questions, in a process that mirrors the hierarchical structure of the grammar. The responses are then processed to produce an action tree.
This allows crowd-sourced workers to provide annotations with no knowledge of the specifics of the grammar described above. For each sentence annotated with the tool, three responses from distinct users were collected, and a sentence was included in the dataset only if at least two out of three responses matched exactly. This yields 1265 annotated prompts, and 817 annotated interactive instructions. A screenshot of the tool is included in Appendix \ref{fig:annotation_task}.

\subsection{Dataset Statistics}
\label{sec:dataset_statistics}

%The CAIP dataset is divided into train, validation, and test sets. The size of each set is shown in Table~\ref{tab:dataset_stats}.
%
%\begin{table}[t]
%\label{tab:dataset_stats}
%\begin{tabular}{l|lll|l}
%Split       & Train   & Val    & Test \\ \midrule % & Total  \\ \hline
%Generated   & 800M(*) & 5K     & 5K   \\ % & 4     \\
%Rephrases   & 30K     & 1K     & 1K  \\ % & 8     \\
%Prompts     & -       &  -     & 1265 \\ %  & 12    \\
%Interactive & -       & -      & 817  \\ %  & 16    \\ \hline
%% Total       & 17      & 18  & 19   & 20
%\end{tabular}
%\caption{Train, validation, and test splits.}
%\end{table}

\paragraph{Action Frequencies} Since the different data collection settings described in Section~\ref{sec:collected_data} imposed different constraints and biases on the crowd-sourced workers, the distribution of actions in each subset of data is therefore very different. For example, in the Interactive Gameplay scenario, workers were given no prior indication of the bot's capabilities, and spent much of their time asking the bot to build things. The action frequencies of each subset are shown in Figure~\ref{fig:action_freqs}.

\paragraph{Grammar coverage} Some crowd-sourced commands describe an action that is outside the scope of the grammar. To account for this, users of the action tree annotation tool are able to mark that a sentence is a command to perform an action that is not listed. The resulting action trees are labelled \textsc{OtherAction}, and their frequency in each dataset in shown in Figure~\ref{fig:action_freqs}. Note that annotators that choose \textsc{OtherAction} still have the option to label other nodes in the action tree like \textsc{location} and \textsc{reference object}.

%% file: modeling.tex
\section{Baseline Models}
\label{sec:modeling}
In order to assess the challenges of the dataset, we implement several baseline models which read a sentence and output an Action Tree, including an adaptation of the Seq2Tree model of \citep{dong2016language} to our grammar.
% In the following Section, $d$ denotes the dimension of the model.

\paragraph{Sentence Encoder} All of our models rely on a sentence encoder. In this work, we use a bidirectional GRU encoder \citep{ChoMGBBSB14} which encodes a sentence of length $T$  $\mathbf{s} = (w_1, \ldots \ w_{T})$ into a sequence of $T$ dimension $d$ vectors:
\begin{equation*}
f_{GRU}(\textbf{s}) = (\mathbf{h}_1,\ldots, \mathbf{h}_T) \in \mathbb{R}^{d \times T}
\end{equation*}

\paragraph{Multi-Headed Attention} Our models also use multi-head attention over the sentence representation. We use the implementation of \cite{opennmt}, with a residual connection. Given $K$ matrices $\textbf{M}^\alpha = (M_1^\alpha,\ldots, M_1^\alpha) \in \mathbb{R}^{d \times d \times K}$, we define:
%  the $K$-headed attention of a query vector $\mathbf{x} \in \mathbb{R}^d$ over a sequence of vectors $(\mathbf{h}_1,\ldots, \mathbf{h}_T) \in \mathbb{R}^{d \times T}$ is computed as:
\begin{align*}
& \alpha^k_n = \text{softmax}\Big(\frac{\mathbf{x}^{\text{T}} M^{\alpha}_k (\mathbf{h}_1,\ldots, \mathbf{h}_T)}{\sqrt{d}} \Big) \\
& \mathbf{x}^{\alpha} = \sum_{k=1}^K {\alpha^k_n}^{\text{T}} (\mathbf{h}_1,\ldots, \mathbf{h}_T) \\
& \text{attn}(\mathbf{x}, (\mathbf{h}_1,\ldots, \mathbf{h}_T); \textbf{M}^\alpha) = \mathbf{x} + \mathbf{x}^{\alpha}
\end{align*}

\subsection{Node Predictions} The output tree is made up of internal, categorical, and span nodes. We denote each of these sets by $\mathcal{I}$, $\mathcal{C}$ and $\mathcal{S}$ respectively, and the full set of nodes as $\mathcal{N} = \mathcal{I} \cup \mathcal{C} \cup \mathcal{S}$. Given a sentence, our aim is to predict the state of each of the nodes $n \in \mathcal{N}$ in the corresponding Action Tree.

Each node in an Action Tree is either active or inactive. We denote the state of a node $n$ by ${a_n \in \{0, 1\}}$. All the descendants of an inactive internal node $n \in \mathcal{I}$ are considered to be inactive.
Additionally, each categorical node $n \in \mathcal{C}$ has a set of possible values $C^n$. Thus, in a specific Action Tree, each active categorical node has a category label ${c_n \in \{1,\ldots,|C^n|\}}$. Finally, active span nodes $n \in \mathcal{S}$ for a sentence of length $T$ have a start and end index ${(s_n, e_n) \in \{1,\ldots,T\}^2}$.

We take the following approach to predicting the state of a tree. First, we compute a node representation $\textbf{r}^n$ for each node $n \in \mathcal{N}$ based on the input sentence $\textbf{s}$:
\begin{align*}
(\textbf{r}_1,\ldots,\textbf{r}_{|\mathcal{N}|}) = f_{REP}((h_1,\ldots, h_T))
\end{align*}
Then, we compute the probabilities of each of the labels as:
\begin{align}
&\forall n \in \mathcal{N}, &p(a_n) &= \sigma(\langle \mathbf{r}_n, \mathbf{p}_n \rangle) \label{eq:int_pred}\\
&\forall n \in \mathcal{C}, &p(c_n) &= \text{softmax}(M^c_n \mathbf{r}_n) \label{eq:cat_pred} \\
&\forall n \in \mathcal{S}, &p(s_n) &= \text{softmax}(\mathbf{r}_n ^{\text{T}} M^s_n (\mathbf{h}_1,\ldots, \mathbf{h}_T)) \nonumber \\
& &p(e_n) &= \text{softmax}(\mathbf{r}_n ^{\text{T}} M^e_n (\mathbf{h}_1,\ldots, \mathbf{h}_T)) \label{eq:span_pred}
\end{align}
where the following are model parameters:
\begin{align*}
\forall n \in \mathcal{N},& \quad \mathbf{p}_n \in \mathbb{R}^d\\
\forall n \in \mathcal{C},& \quad M^c_n \in \mathbb{R}^{d \times d} \\
\forall n \in \mathcal{S},& \quad (M^s_n, M^e_n) _n \in \mathbb{R}^{d \times d \times 2}
\end{align*}
Our proposed baselines differ from each other by how we compute the node representations $\mathbf{r}_n$ from the sentence. We present three implementations $f_{REP}$ in Section~\ref{sec:node_rep}.

\subsection{Node Representation}
\label{sec:node_rep}

\paragraph{Independent predictions} Our first model computes $\mathbf{r}_n$ independently for each node by attending over the sentence representation. More specifically, each node $n \in \mathcal{N}$ has a parameter $\mathbf{v}_n \in \mathbb{R}^n$. We compute $\mathbf{r}_n$ by simply using $\mathbf{v}_n$ to attend over the sequence encoding $(\mathbf{h}_1,\ldots,\mathbf{h}_T)$ using $K$ headed attention parameterized $\textbf{M}^{\nu} \in \mathbb{R}^{d \times d \times K}$:
\begin{align}
\mathbf{r}_n &= \text{attn}(\mathbf{x}, (\mathbf{h}_1,\ldots, \mathbf{h}_T); \textbf{M}^\nu)
\end{align}

\paragraph{Seq2Tree} We also implement the recurrent node representation function from Seq2Tree model of \citep{dong2016language}. It uses a recurrent decoder to compute representations for the children of a node in sequence based on the previously predicted siblings and the parent's representation. So let $n^p \in \mathcal{I}$ be an internal node, let $(c_1, \ldots, c_m)$ be its children. Let the recurrent hidden state of a node $n$ be noted as $\mathbf{g}^n$, and $\circ$ be the concatenation. We then compute:
\begin{align}
& \mathbf{r}_{c_t} = \text{attn}(\mathbf{v}_{c_t} + \mathbf{g}^{c_{t-1}}, (\mathbf{h}_1,\ldots, \mathbf{h}_T); \textbf{M}^\sigma) \\
& \mathbf{g}^{c_{t}} =
\begin{cases}
    f_{rec}(\mathbf{g}^{c_{t-1}}, \mathbf{v'}_{c_t} \circ \mathbf{g}^{n^p}), & \text{if } a_{c_t} = 1\\
    \mathbf{g}^{c_{t-1}}, & \text{else}
\end{cases}
\end{align}
Where $\textbf{M}^{\nu} \in \mathbb{R}^{d \times d \times K}$ is a tree-wise parameter (as in the independent prediction case), $f_{rec}$ is the GRU recurrence function, and $\mathbf{v'}_{c_t}$ is a node parameter (one per category for categorical nodes).

\paragraph{SentenceRec} One possible limitation of the Seq2tree model predicted above is that the tree-side recurrent update do not directly depend on the input sentence. This can be addressed by a simple modification: we simply add the node representation $\mathbf{r}_{c_t}$ to the input for the recurrent update:
\begin{align}
& \mathbf{r}_{c_t} = \text{attn}(\mathbf{v}_{c_t} + \mathbf{g}^{c_{t-1}}, (\mathbf{h}_1,\ldots, \mathbf{h}_T); \textbf{M}^\sigma) \\
& \mathbf{g}^{c_{t}} =
\begin{cases}
    f_{rec}(\mathbf{g}^{c_{t-1}}, (\mathbf{v'}_{c_t} + \mathbf{r}_{c_t}) \circ \mathbf{g}^{n^p}), & \text{if } a_{c_t} = 1\\
    \mathbf{g}^{c_{t-1}}, & \text{else}
\end{cases}
\end{align}
We refer to this model as SentenceRec.

\subsection{Sequential Prediction}

We predict the sate of the Action Tree given a sentence in a sequential manner, by predicting the state of the nodes ($\{a_n; \; \forall n \in \mathcal{N}\}$, $\{c_n; \; \forall n \in \mathcal{C}\}$, and $\{(s_n, e_n); \; \forall n \in \mathcal{S}\}$) in Depth First Search order. Additionally, since an inactive node's descendant are all inactive, we can skip the sub-trees rooted at $n$ if we predict $a_n=0$. Let us thus note the parent of a node $n$ as $\pi(n)$. Given Equations~\ref{eq:int_pred} to \ref{eq:span_pred}, the log-likelihood of a tree with states $(\textbf{a}, \textbf{c}, \textbf{s}, \textbf{e})$ given a sentence $\textbf{s}$ can be written as:
\begin{align}
\mathcal{L} & = \sum_{n \in \mathcal{N}} a_{\pi(n)} \log(p(a_n)) \nonumber \\
& \quad + \sum_{n \in \mathcal{C}} a_n \log(p(c_n)) \nonumber \\
& \quad + \sum_{n \in \mathcal{S}} a_n \Big(\log(p(s_n)) + \log(p(e_n))\Big)
\end{align}
Not that since in all of our models the representation $\textbf{r}_n$ of a node $n$ only depends on nodes that have been seen before it in a DFS search, this loss lends itself well to beam search prediction.

%% file: experiments.tex
\section{Experiments}
\label{sec:experiments}

\begin{table*}
\small
\center
\begin{tabular}{llccccc}
Sampling     &   Model          & Temp.  Valid.         & Rep.  Valid.         & Rep.   Test  & Prompts  & Interactive \\
\midrule
\multirow{3}{*}{Rephrases}       & Independent & 0.979          & 0.810          & \textbf{0.806}     & 0.171            & 0.307     \\
                                 & Seq2Tree    & 0.979          & 0.801          & 0.794     & 0.180            & 0.231     \\
                                 & SentenceRec & 0.976          & 0.814          & \textbf{0.807}     & 0.159            & 0.255     \\
\midrule
\multirow{3}{*}{Prompts}         & Independent & 0.976          & 0.804          & 0.773     & 0.184            & 0.370     \\
                                 & Seq2Tree    & 0.987          & 0.819          & 0.789     & 0.176            & 0.321     \\
                                 & SentenceRec & 0.980          & 0.828          & 0.776     & 0.179            & 0.360     \\
\midrule
\multirow{3}{*}{Interactive}     & Independent & 0.976          & 0.782          & 0.709     & 0.179            & 0.337     \\
                                 & Seq2Tree    & 0.975          & 0.802          & 0.734     & \textbf{0.196}            & 0.454     \\
                                 & SentenceRec & 0.980          & 0.820          & 0.771     & \textbf{0.195}            & \textbf{0.465}    
\end{tabular}
\caption{Success of trained models over various training and test distributions.  Each group of three rows corresponds to a distribution over top-level commands used during training. 
``Rephrases'', ``Prompts'', and ``Interactive'' as in Figure \ref{fig:action_freqs}.  In the columns, ``Temp'' refers to the templates distribution, ``Rep.'' to rephrases (from the template distribution), and ``Prompts'' and ``Interactive'' as before.\label{tab:accuracies}
}
\end{table*}

\begin{table}
\small
\begin{tabular}{llccc}
                  Model                     & Node & \multicolumn{3}{c}{Test P/R/F}   \\
                                         &      & Rephrases & Prompts & Interactive    \\
%\midrule
%\multirow{6}{*}{R} & \multirow{3}{*}{Ind.} & INT  & 99/97/98  & 75/72/73  & 85/71/78 \\
%                   &                       & CAT  & 96/94/97  & 43/52/47  & 57/59/58 \\
%                   &                       & SPAN & 95/93/94  & 50/34/41  & 58/29/39 \\
%\cmidrule{2-6}
%                   & \multirow{3}{*}{SRec} & INT  & 98/97/97  & 68/75/72  & 69/71/70 \\
%                   &                       & CAT  & 94/93/94  & 38/51/43  & 46/55/50 \\
%                   &                       & SPAN & 94/93/94  & 40/37/38  & 32/26/29 \\
\midrule
 \multirow{3}{*}{Ind.} & INT  & 97/95/96  & 75/77/76  & 75/83/79 \\
                    & CAT  & 92/90/91  & 44/56/50  & 52/66/58 \\
                   & SPAN & 94/91/93  & 53/42/47  & 58/51/54 \\
\midrule
   \multirow{3}{*}{SRec} & INT  & 97/96/97  & 75/77/76  & 86/84/85 \\
                    & CAT  & 92/91/92  & 42/54/47  & 59/66/62 \\
                   & SPAN & 94/93/94  & 51/43/46  & 68/56/61
\end{tabular}
\caption{Per-node Precision, Recall and F1 for models trained with interactive sampling for all node types  \label{tab:per-node}
}
\end{table}

% \subsection{Training Setup}

\paragraph{Training Data} We train our model jointly on the (virtually unlimited) template generations and set of 33K training rephrases. Early experiments showed that a model trained exclusively on templated generations failed to reach accuracies better than 40\% on the validation rephrases. Training on rephrases did a little better (up to 65\%) but still trailed behind models trained on both (around 80\%, see Tale~\ref{tab:accuracies}).

The action types represented in all three test datasets (rephrases, prompts and interactive) are very different, as shown in Figure~\ref{fig:action_freqs}. In order to address both of these issues, we sample training examples evenly between templates and rephrases according to each of the test setting distributions (no replacement til all examples of a subset have been seen).
%  The choice of the distribution has a significant effect on the model behavior, as shown next.

\paragraph{Modeling Details}  We use a 2-layer GRU sentence encoder and all hidden layers in our model have dimension $d=256$. We use pre-trained word embeddings computed with FastText with subword information \citep{BojanowskiGJM17}, to which we concatenate free learnable dimensions (these are initialized to be 0, and we tried adding 0, 8, 32 and 64 free dimensions). All models are trained with Adagrad, using label smoothing, dropout, and word dropout for regularization. In all settings, we selected the model which reached the best accuracy on the validation rephrases to evaluate on the test sets. We provide our model and training code.

% \subsection{Results}

\paragraph{Overview of Results} Table~\ref{tab:accuracies} presents tree-level accuracies for the proposed training settings. First, we notice that all models are able to reach near-perfect accuracy on generations from our templates, which means they can invert the generation process described in Section~\ref{sec:generated_data}. The accuracy on the validation and test rephrased data is also high, up to 80.7\% for the SentenceRec model. However, the worse performance on instructions from both prompts and interactive shows that our setting poses significant generalization challenges. In particulars, all models have significant trouble with the prompts, which come from crowd-sourced workers asked to imagine general game commands and may not fit the exact Minecraft setting. Still, 86\% of the annotations are valid under our grammar, and we hope that future work will be better able to address the domain shift to be able to predict those.

On the ``interactive'' commands , the models do a little better. In general, the SentenceRec seems to have a small edge over the base Seq2Tree model, but the main difference seems to be between the independent prediction and recurrent models. While the latter do much better when trained in-distribution (12\% absolute gap), the former does seem to adapt better to the distribution shift when trained using the rephrases or prompts sampling.

\paragraph{Analysis} Table~\ref{tab:per-node} gives insights into model behaviors on CATegorical, INTernal and SPAN nodes. Accurate prediction of a categorical or span node depends on having predicted all of the internal nodes on the path to the root, which explains why CAT and SPAN P/R/F numbers are lower than INT. Additionally, both models have more trouble predicting span than categorical nodes.

We also computed confusion matrices for the best SentenceRec model (see Appendix~\ref{sec:conf_mat}). For internal nodes, both models seem to have trouble identifying the scope of some \textsc{location} and \textsc{repeat} nodes: i.e. even when identifying that the command specifies a location, is it the location where the command needs to be executed, or the action where the command's argument is located? There is also a confusion between \textsc{schematic} and \textsc{action reference objects}, which we assume comes from the difficulty of interpreting whether the speaker is asking the model to build an object it knows (\textsc{schematic}) or another copy of an object in the world (\textsc{reference object}), a prediction which must rely on an understanding of the context.

Finally, the internal parent being absent seems to account for most of the CAT and SPAN mistakes (aside from the action type, which is a child of the root). For action types, the model seems to have trouble recognizing questions and Fill requests mostly. The model also seems to often confuse Mobs (animated creatures in the game) with Objects, which is indeed difficult to disambiguate without some background knowledge. For spans, the model mostly makes the mistake of predicting a node as inactive when it is present. It should be noted that span mis-match are especially rare, except for the model sometimes confusing the depth and height of an object when both are present.         

%\begin{table*}
%\small
%\resizebox{\columnwidth}{!}{
%\begin{tabular}{l | c | c | c }
%               & Indep & Seq2tree & SenRec \\ 
%\hline\hline
%S + T & .981               & .984        & .982       \\
%S + R & .808              & .809        & .817        \\
% R      & .81                & .79           & .81         \\
% HB   & .31                & .23           &.25     
%\end{tabular}  
%  }
% \resizebox{\columnwidth}{!}{
%\begin{tabular}{l | c | c | c }
%                                                                                              & independent & seq2tree & sen2tree \\ 
%\hline\hline                                                                                              
%S + T & .976               & .981        & .98       \\
%S + R & .786              & .816        & .825        \\
%R        & .71                & .73           & .76         \\
%HB   & .34                & .45           &.43     
% \end{tabular*}
%}      
%\caption{Success of trained models over various test and train distributions.   Each row corresponds in the two tables corresponds to a test distribution.  
%S + T  and S + R is for templates and rephrases respectively sampled with the top-level action distribution of human-bot dialogues \ref{sec:data}.  R is for rephrases sampled from the template-generation distribution, and HB is human-bot dialogue data.  The top table uses the natural template distribution for training, and the bottom one uses the human-bot top-level distribution. }
%\end{table}

%% file: action_tree.tex
\onecolumn

\section{Action Tree structure}
\label{sec:action_tree}

\begin{figure*}[ht!]
    \centering
    \includegraphics[width=\linewidth]{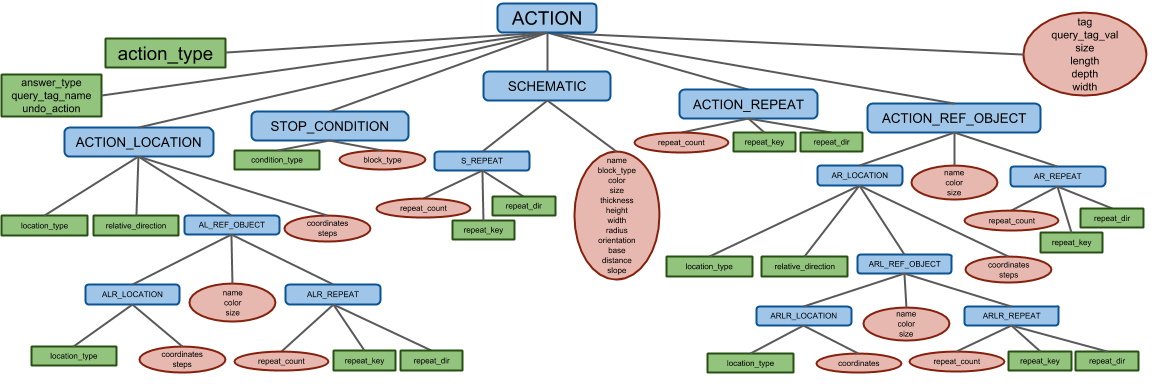}
    \caption{Action space grammar}
    \label{fig:action_grammar}
\end{figure*}

This section describes the details of the action tree.
We support the following actions in our dataset : Build, Copy, Noop, Spawn, Resume, Fill, Destroy, Move, Undo, Stop, Dig, Tag, FreeBuild and Answer.
The detailed action tree for each action has been presented in the following subsections. Figure~\ref{fig:action_tree_ex} shows an example for a \textsc{build} action.

\begin{figure}[ht]
    \centering
    \small
    \begin{verbatim}
  0     1    2   3     4    5   6
"Make three oak wood houses to the
 7   8   9   10   11    12
left of the dark grey church."

{"Build": {
  "schematic": {
   "has_block_type_": [2, 3],
   "has_name_": [4, 4],
   "repeat": {
    "repeat_key": "FOR",
    "repeat_count": [1, 1]
  }},
  "location": {
   "relative_direction": "LEFT",
   "location_type": "BlockObject",
   "location_reference_object": {
    "has_colour_": [10, 11],
    "has_name_": [12, 12]
}}}}
    \end{verbatim}
    \vspace{-20pt}
    \caption{An example action tree. The word indices are numbered here for clarity.}
    \vspace{-8pt}
    \label{fig:action_tree_ex}
\end{figure}

\subsection{ Build Action}
\begin{figure}[h]
    \centering
    \includegraphics[width=15cm,height=20cm,keepaspectratio]{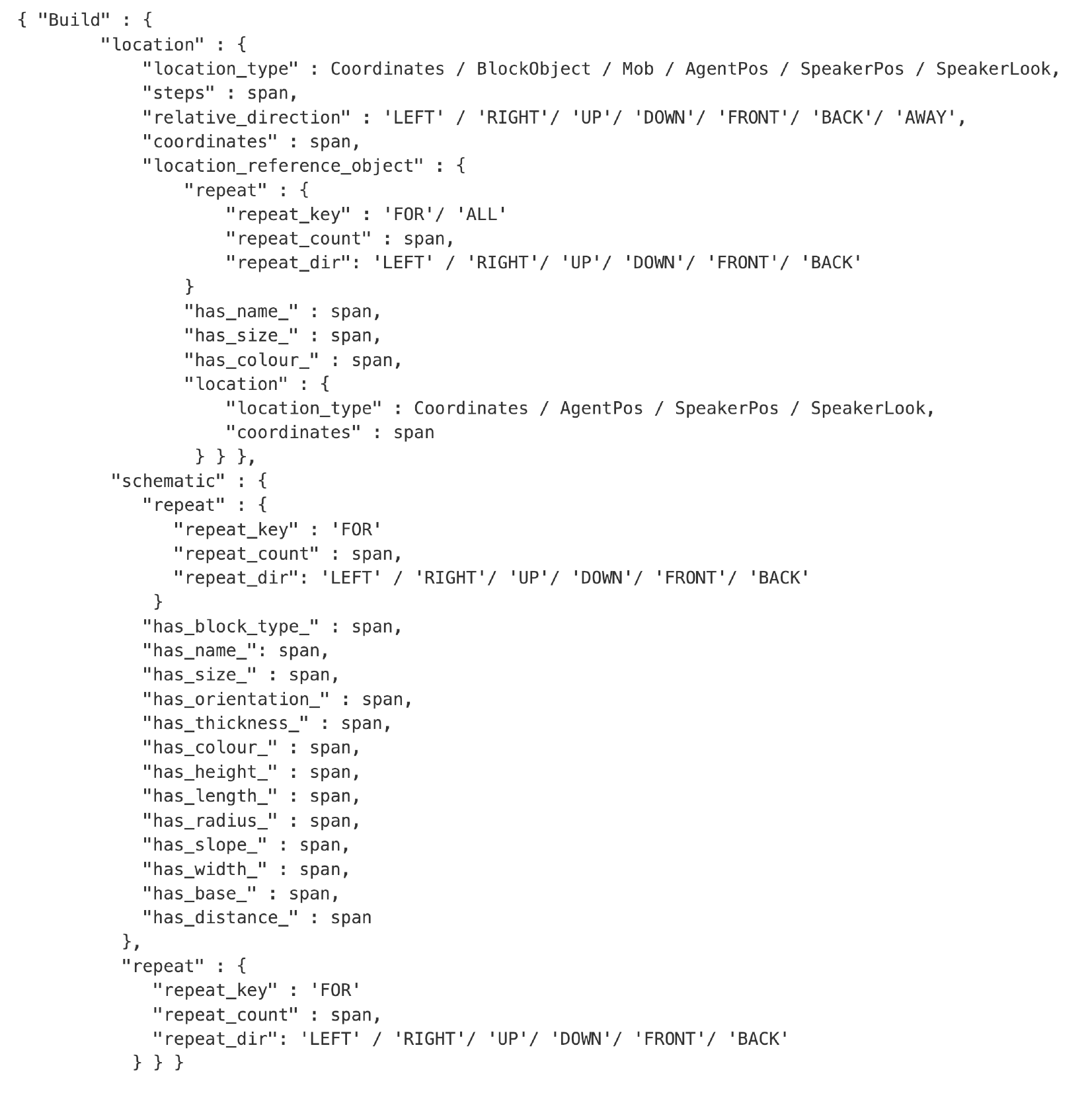}
    \caption{Details of Build action tree}
    \label{fig:build_dict}
\end{figure}

This is the action to Build a schematic at an optional location. The Build action tree is shown in \ref{fig:build_dict} and the action can have one of the following as its child:
\begin{itemize}
	\setlength\itemsep{0.0em}
	\item location only
	\item schematic only
	\item location and schematic both
	\item neither
\end{itemize}

\subsection{Copy Action}
\begin{figure}[h]
    \centering
    \includegraphics[width=15cm,height=20cm,keepaspectratio]{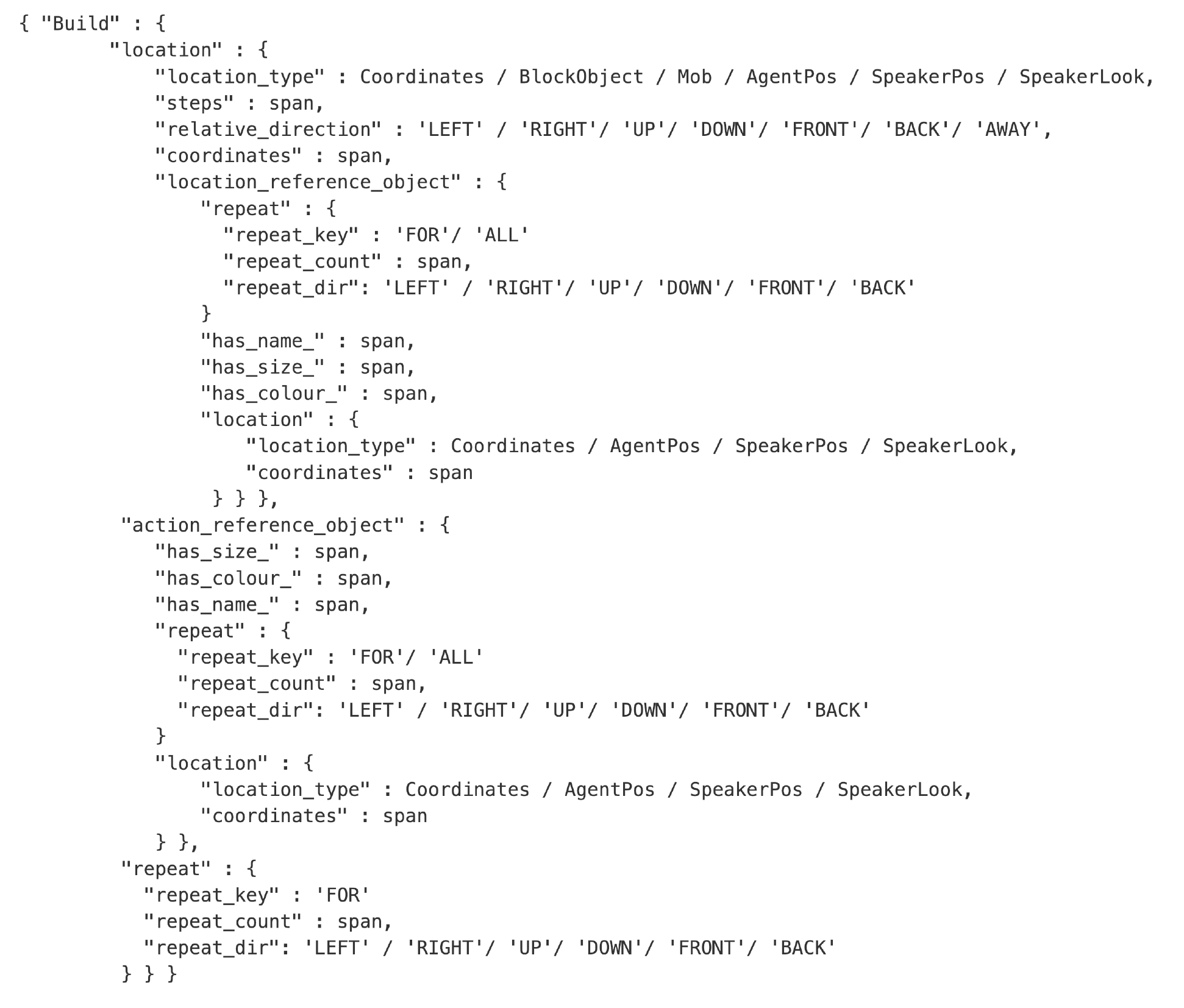}
    \caption{Details of Copy action tree}
    \label{fig:copy_dict}
\end{figure}

This is the action to copy a block object to an optional location. The copy action is represented as a "Build" with an optional "action reference object" in the tree. The tree is shown in \ref{fig:copy_dict}.

Copy action can have one the following as its child:
\begin{itemize}
	\setlength\itemsep{0.0em}
	\item action reference object
	\item action reference object and location
	\item neither
\end{itemize}

\subsection{ Spawn Action}
\begin{figure}[h]
    \centering
    \includegraphics[width=10cm,height=10cm,keepaspectratio]{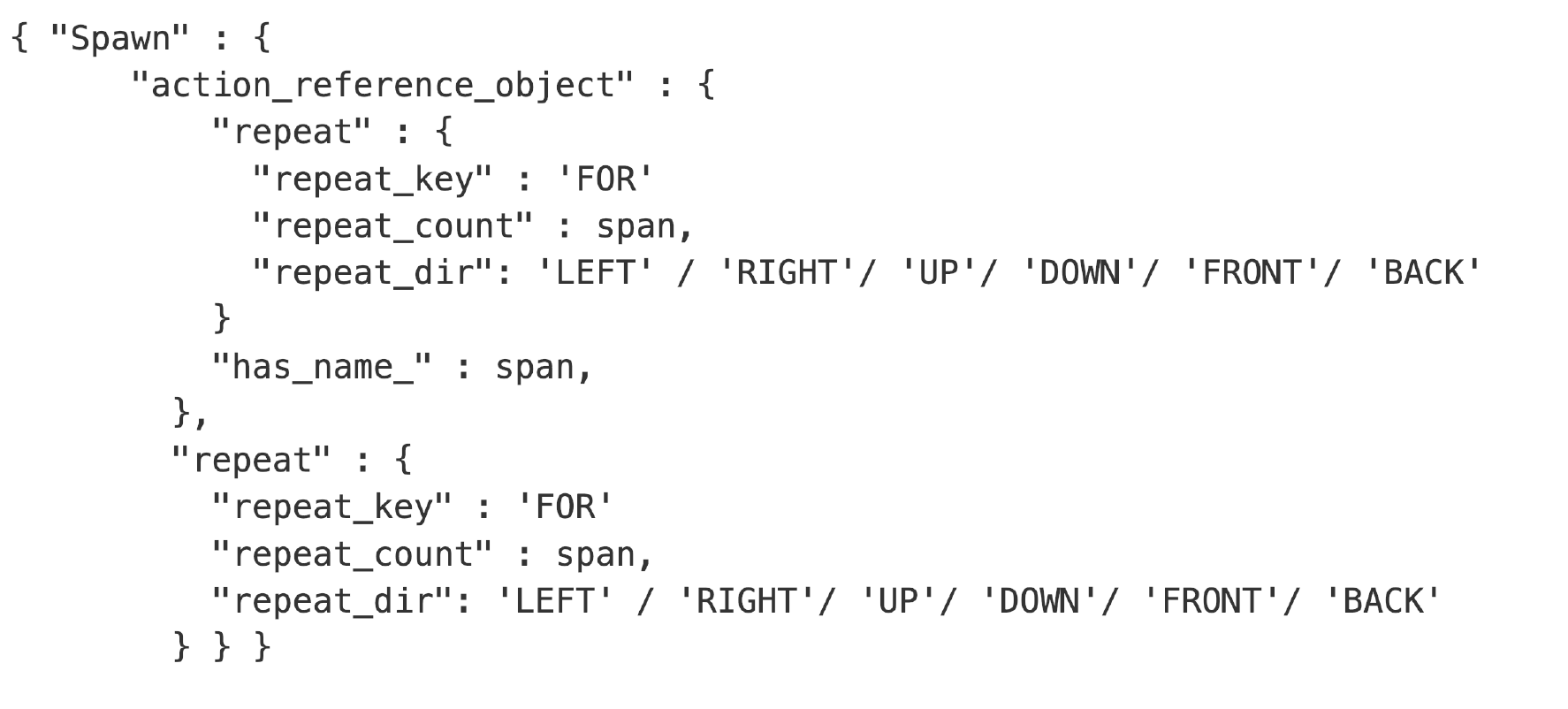}
    \caption{Details of Spawn action tree}
    \label{fig:spawn_dict}
\end{figure}

This action indicates that the specified object should be spawned in the environment. The tree is shown in: \ref{fig:spawn_dict}

Spawn action has the child: action reference object.

\subsection{ Fill Action}
\begin{figure}[h]
    \centering
    \includegraphics[width=15cm,height=20cm,keepaspectratio]{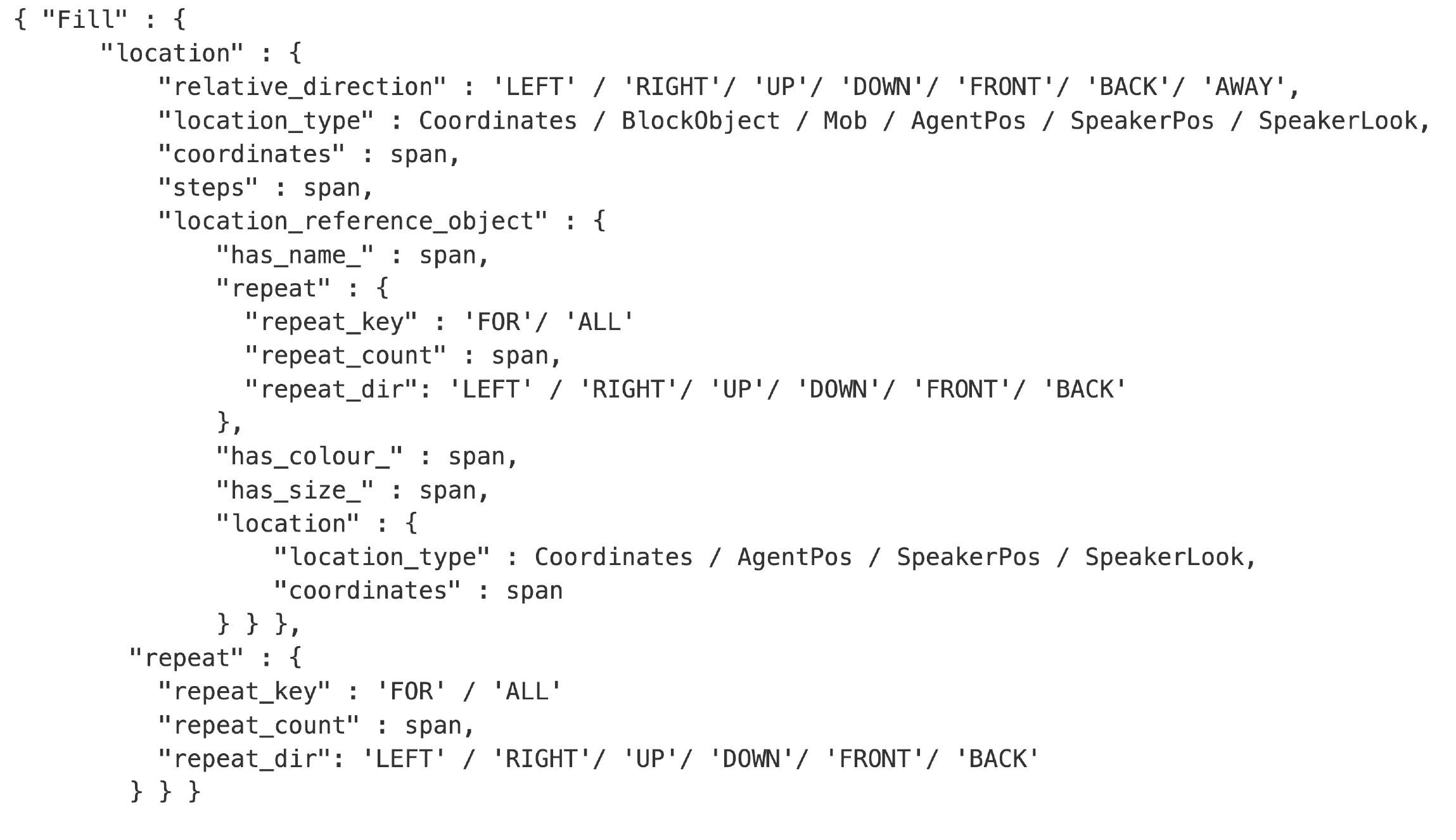}
    \caption{Details of Fill action tree}
    \label{fig:fill_dict}
\end{figure}

This action states that a hole / negative shape at an optional location needs to be filled up. The tree is explained in : \ref{fig:fill_dict}

Fill action can have one of the following as its child:
\begin{itemize}
	\setlength\itemsep{0.0em}
	\item location
	\item nothing
\end{itemize}

\subsection{ Destroy Action}
\begin{figure}[h]
    \centering
    \includegraphics[width=15cm,height=20cm,keepaspectratio]{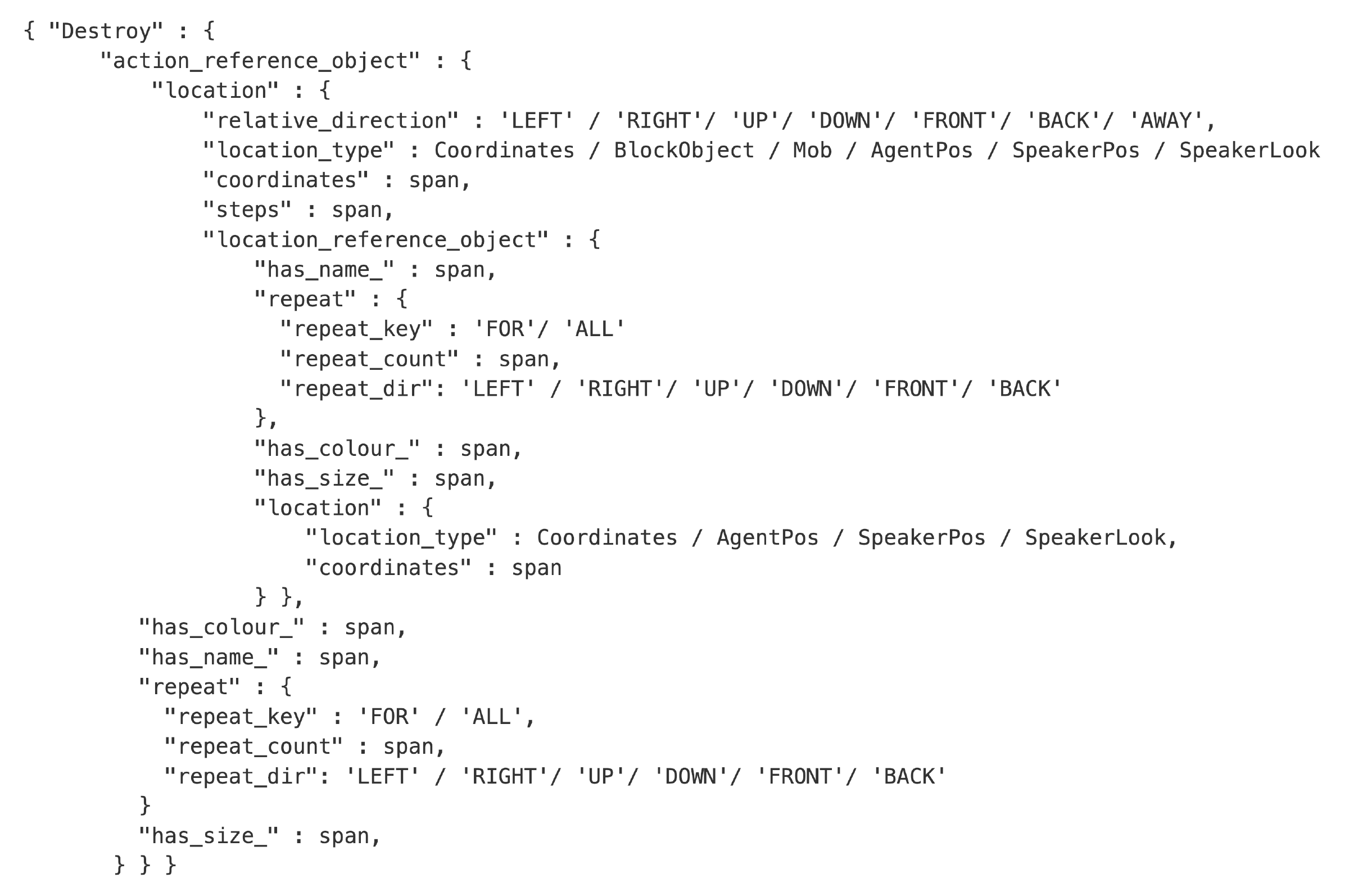}
    \caption{Details of Destroy action tree}
    \label{fig:destroy_dict}
\end{figure}

This action indicates the intent to destroy a block object at an optional location. The tree is shown in: \ref{fig:destroy_dict}

Destroy action can have one of the following as the child:
\begin{itemize}
	\setlength\itemsep{0.0em}
	\item action reference object
	\item nothing
\end{itemize}

\subsection{Move Action}
\begin{figure}[h]
    \centering
    \includegraphics[width=15cm,height=20cm,keepaspectratio]{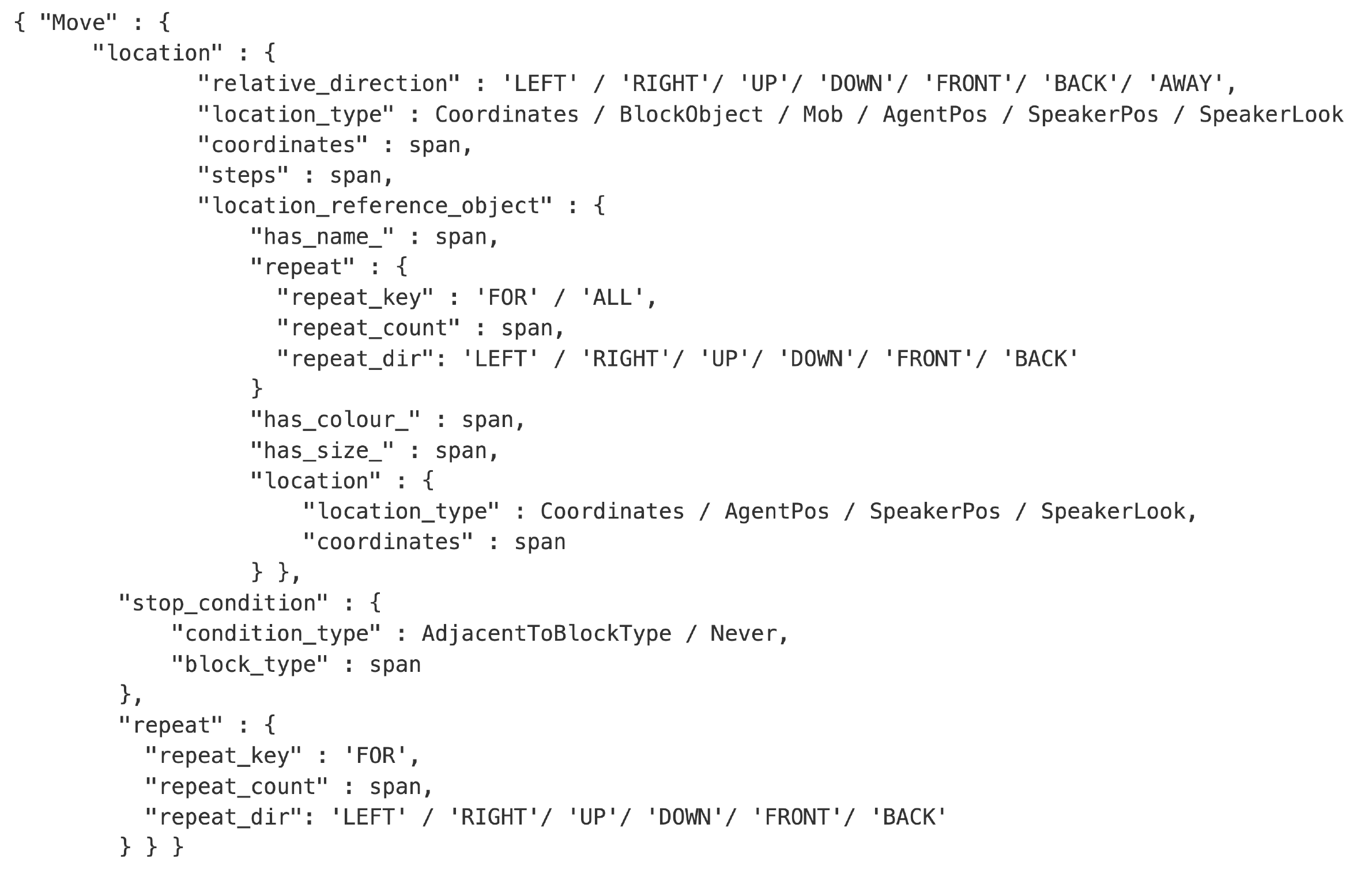}
    \caption{Details of Move action tree}
    \label{fig:move_dict}
\end{figure}

This action states that the agent should move to the specified location, the corresponding tree is in: \ref{fig:move_dict}

Move action can have one of the following as its child:
\begin{itemize}
	\setlength\itemsep{0.0em}
	\item location
	\item stop condition (stop moving when a condition is met)
	\item location and stop condition
	\item neither
\end{itemize}

\subsection{ Dig Action}
\begin{figure}[h]
    \centering
    \includegraphics[width=15cm,height=20cm,keepaspectratio]{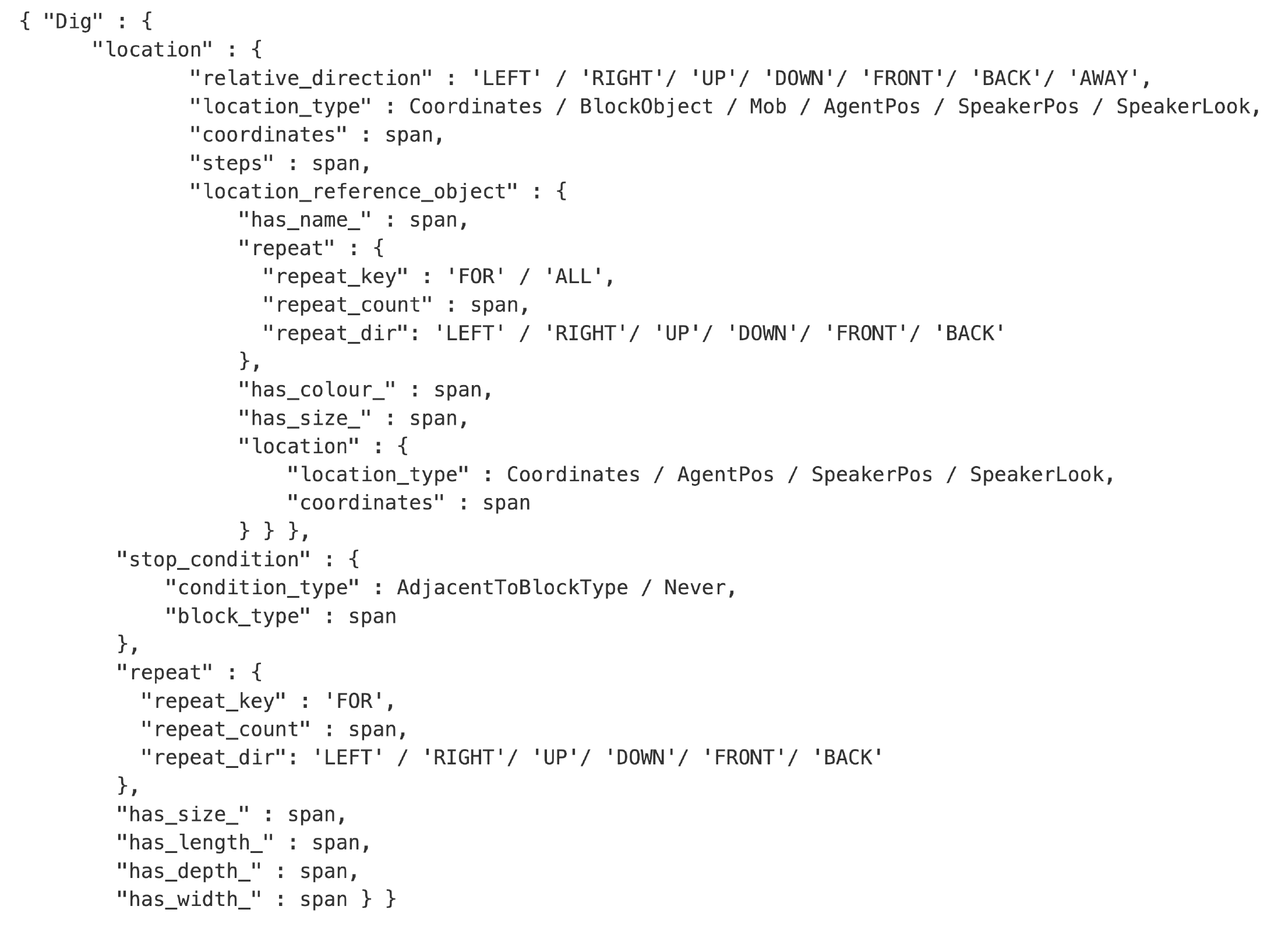}
    \caption{Details of Dig action tree}
    \label{fig:dig_dict}
\end{figure}

This action represents the intent to dig a hole / negative shape of optional dimensions at an optional location. The tree is in \ref{fig:dig_dict}

Dig action can have one of the following as its child:
\begin{itemize}
	\setlength\itemsep{0.0em}
	\item nothing
	\item location
	\item stop condition 
	\item location and stop condition and / or size, length, depth, width
\end{itemize}

\subsection{Tag Action}
\begin{figure}[h]
    \centering
    \includegraphics[width=10.8cm,height=15cm,keepaspectratio]{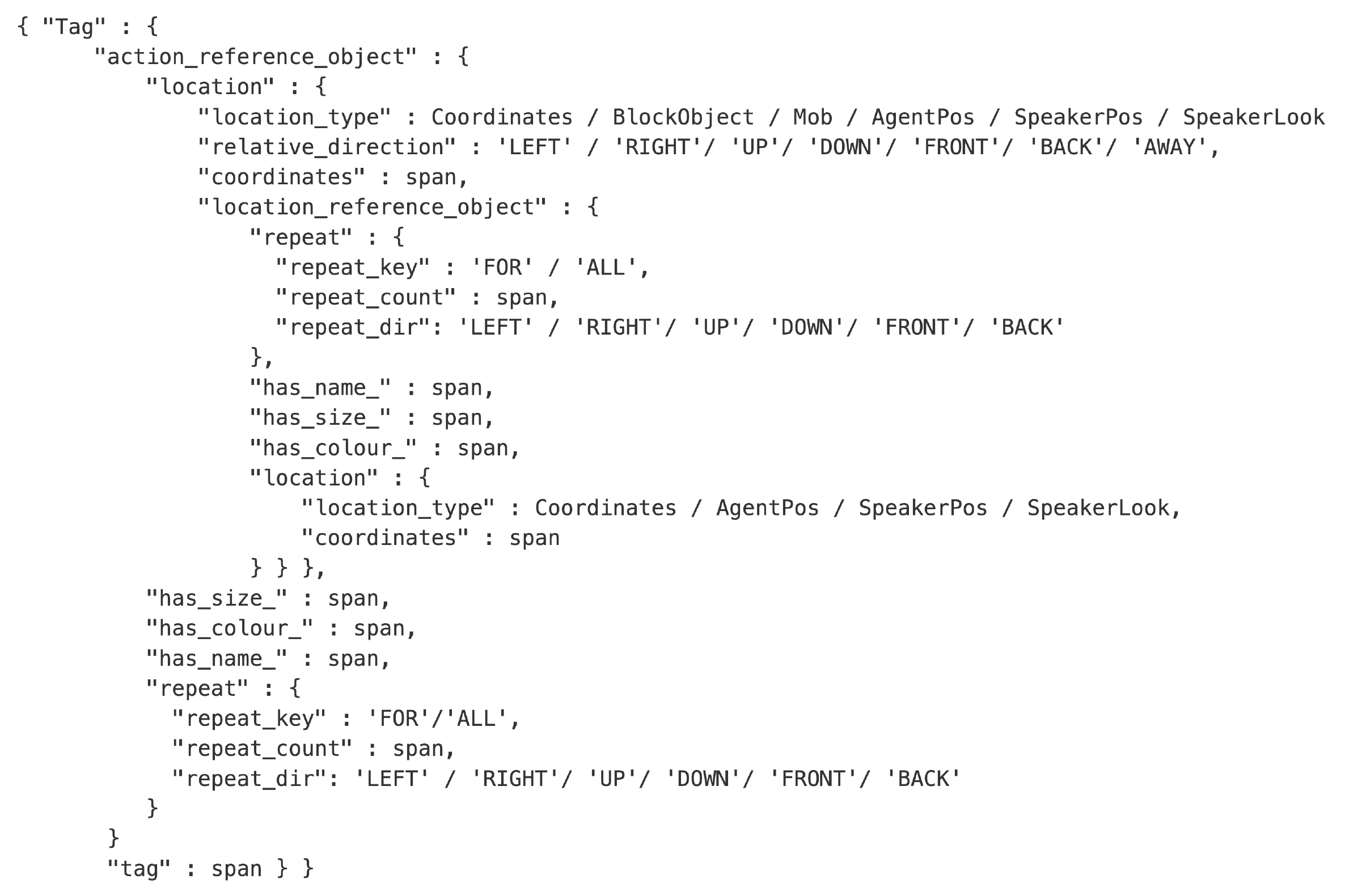}
    \caption{Details of Tag action tree}
    \label{fig:tag_dict}
\end{figure}

This action represents that an action reference object should be tagged with the given tag and the tree is shown in: \ref{fig:tag_dict}

Tag action can have the following as its children:
\begin{itemize}
	\setlength\itemsep{0.0em}
	\item tag
	\item action reference object
\end{itemize}

\subsection{FreeBuild Action}
\begin{figure}[h]
    \centering
    \includegraphics[width=15cm,height=20cm,keepaspectratio]{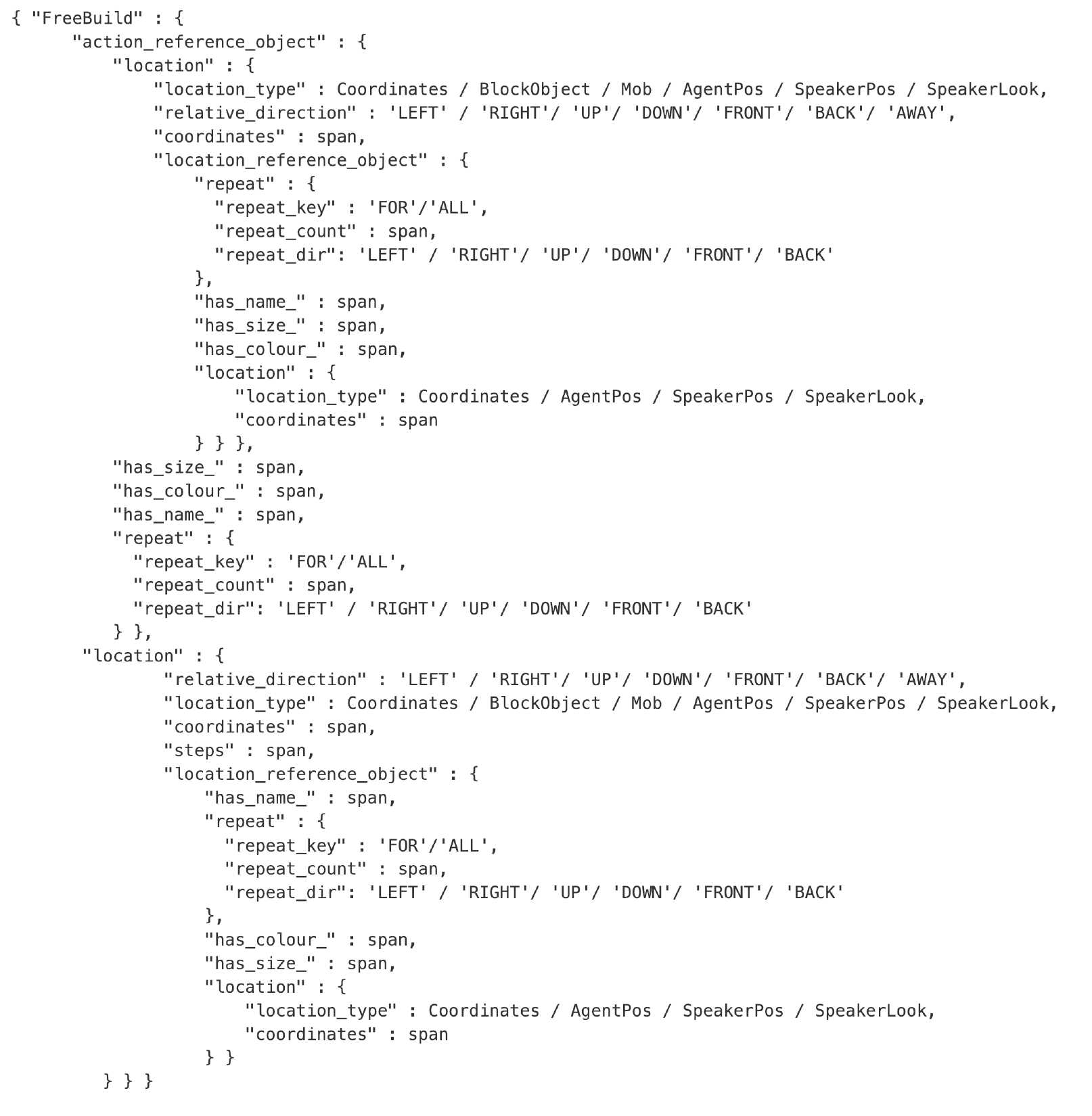}
    \caption{Details of FreeBuild action tree}
    \label{fig:freebuild_dict}
\end{figure}

This action represents that the agent should complete an already existing half-finished block object, using its mental model. The action tree is explained in: \ref{fig:freebuild_dict}

FreeBuild action can have one of the following as its child:
\begin{itemize}
	\setlength\itemsep{0.0em}
	\item action reference object only
	\item action reference object and location
\end{itemize}

\subsection{Answer Action}
\begin{figure}[h]
    \centering
    \includegraphics[width=12cm,height=12cm,keepaspectratio]{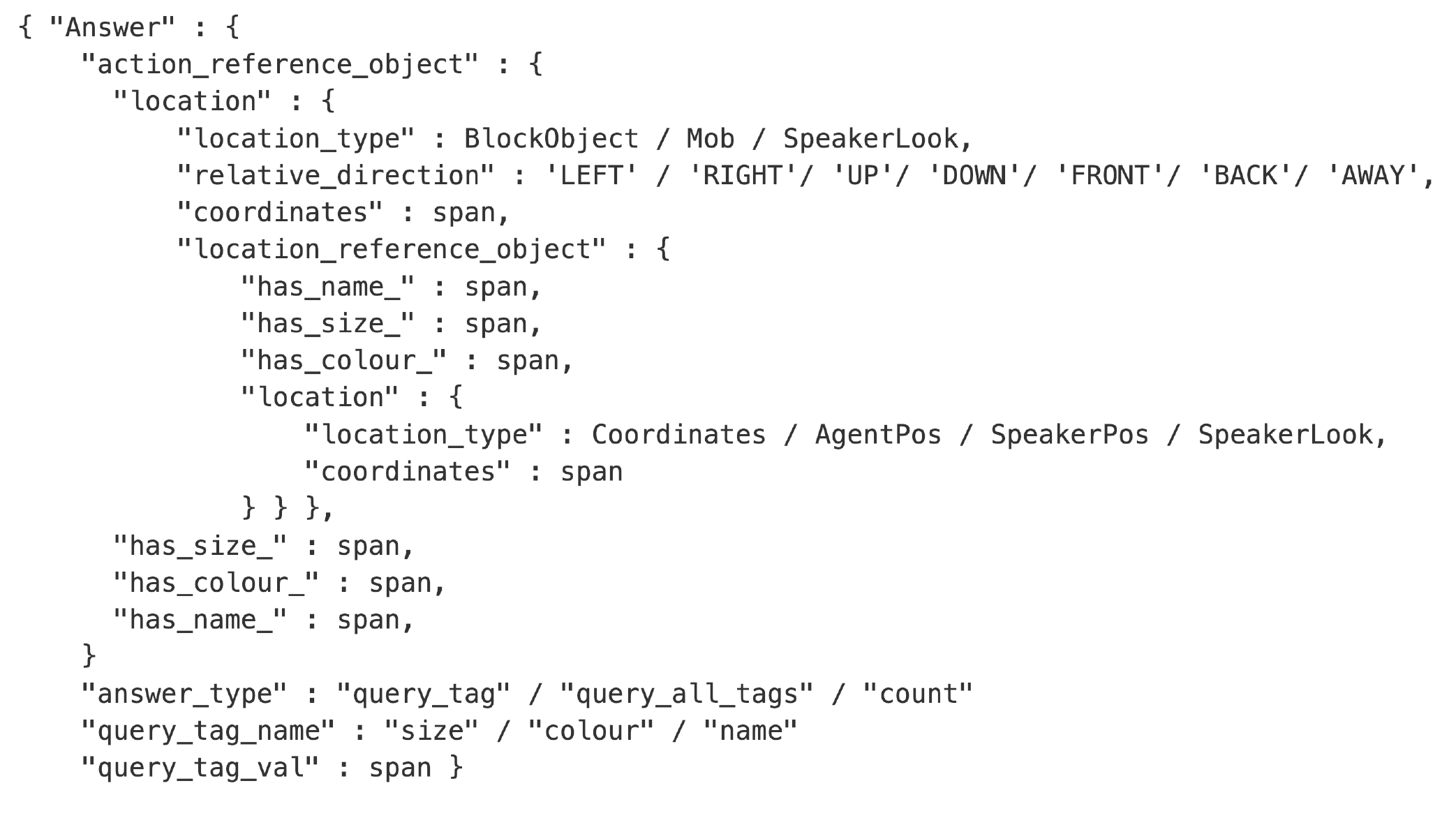}
    \caption{Details of Answer action tree}
    \label{fig:answer_dict}
\end{figure}

This action represents the agent answering a question about the environment.
This is similar to the setup in Visual Question Answering. The tree is represented in: \ref{fig:answer_dict}

Answer action has the following as its children: action reference object, answer type, query tag name and query tag val.
Answer type represents the type of expected answer : counting, querying a specific attribute or querying everything ("what is the size of X" vs "what is X" )

\subsection{ Noop Action}
\begin{figure}[h]
	\centering
    \includegraphics[width=5cm,height=5cm,keepaspectratio]{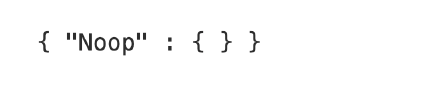}
    \caption{Details of Noop action tree}
    \label{fig:noop_dict}
\end{figure}

This action indicates no operation should be performed, the tree is shown in : \ref{fig:noop_dict}

\subsection{Resume Action}
\begin{figure}[h]
    \centering
    \includegraphics[width=5cm,height=5cm,keepaspectratio]{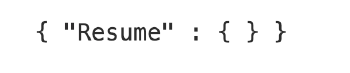}
    \caption{Details of Resume action tree}
    \label{fig:resume_dict}
\end{figure}

This action indicates that the previous action should be resumed, the tree is shown in: \ref{fig:resume_dict}

\subsection{ Undo Action}
\begin{figure}[h]
    \centering
    \includegraphics[width=5cm,height=5cm,keepaspectratio]{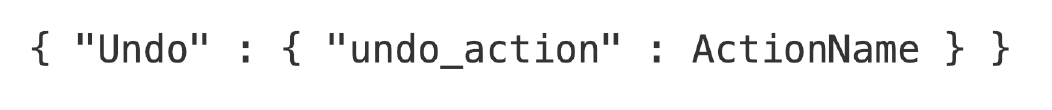}
    \caption{Details of Undo action tree}
    \label{fig:undo_dict}
\end{figure}

This action states the intent to revert the specified action, if any. The tree is in \ref{fig:undo_dict}.
Undo action can have on of the following as its child:
\begin{itemize}
	\setlength\itemsep{0.0em}
	\item undo action
	\item nothing (meaning : undo the last action)
\end{itemize}

\subsection{ Stop Action}
\begin{figure}[h]
    \centering
    \includegraphics[width=5cm,height=5cm,keepaspectratio]{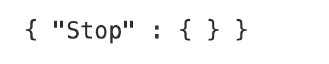}
    \caption{Details of Stop action tree}
    \label{fig:stop_dict}
\end{figure}

This action indicates stop and the tree is shown in \ref{fig:stop_dict}

%% file: task_instructions.tex
\clearpage
\section{Crowd-sourced task instructions}
We have listed the instructions for each task mentioned in section \ref{sec:collected_data} in the following subsections.

\subsection{Image and Text Prompts}
\label{sec:freegen_instructions}
\begin{figure}
	\includegraphics[width=\linewidth ]{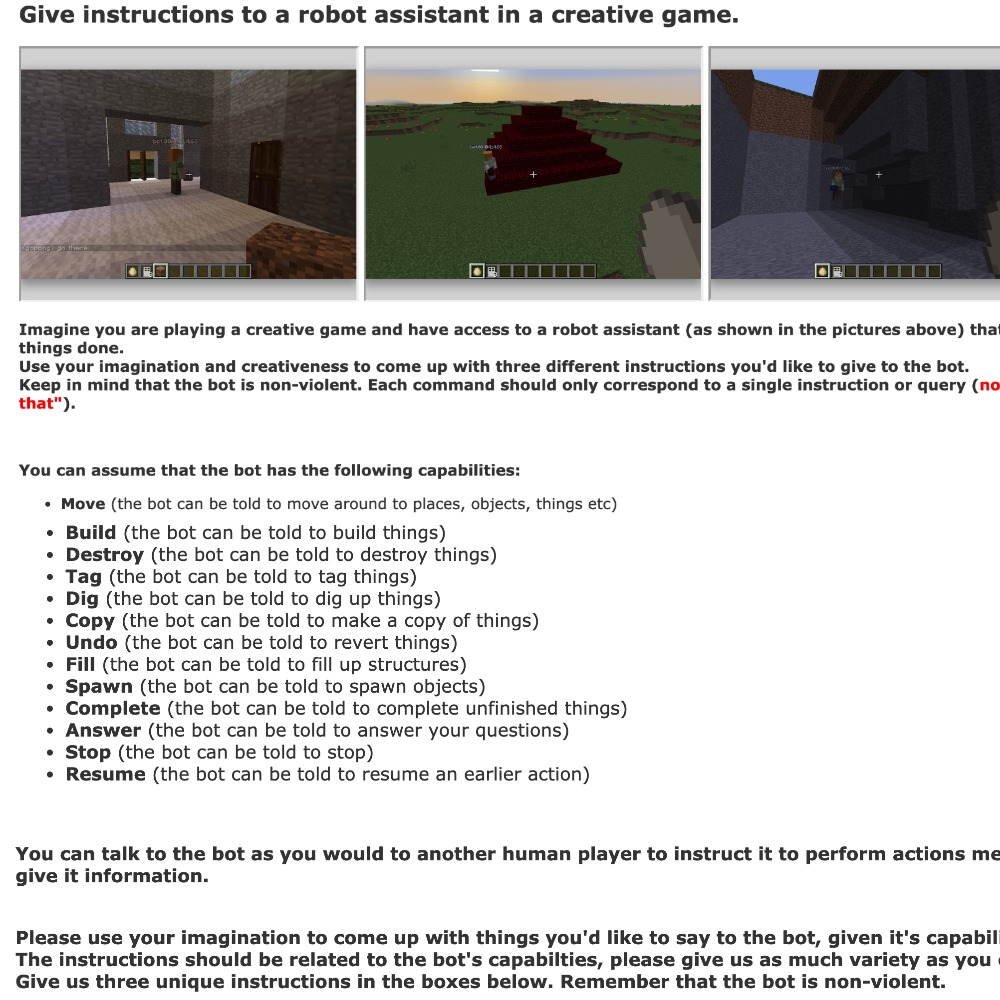}
	\caption{The task instructions shown to crowd-sourced workers for the Image and text prompts task\label{fig:freegen}}
\end{figure}

The instructions shown to workers are shown in \ref{fig:freegen}.

\subsection{Interactive Gameplay}
\label{sec:appen}
\begin{figure}
	\includegraphics[width=\linewidth ]{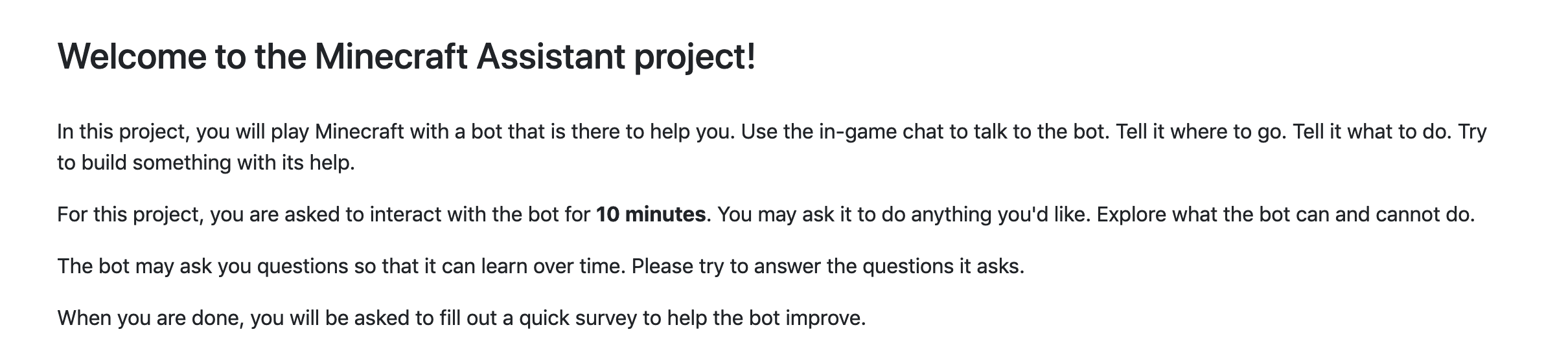}
	\caption{The task instructions shown to crowd-sourced workers for the interactive game play\label{fig:appen}}
\end{figure}

The instructions shown to workers are shown in \ref{fig:appen}.

\subsection{Annotation tool}
\label{sec:tool_instructions}
\begin{figure}
	\includegraphics[width=\linewidth ]{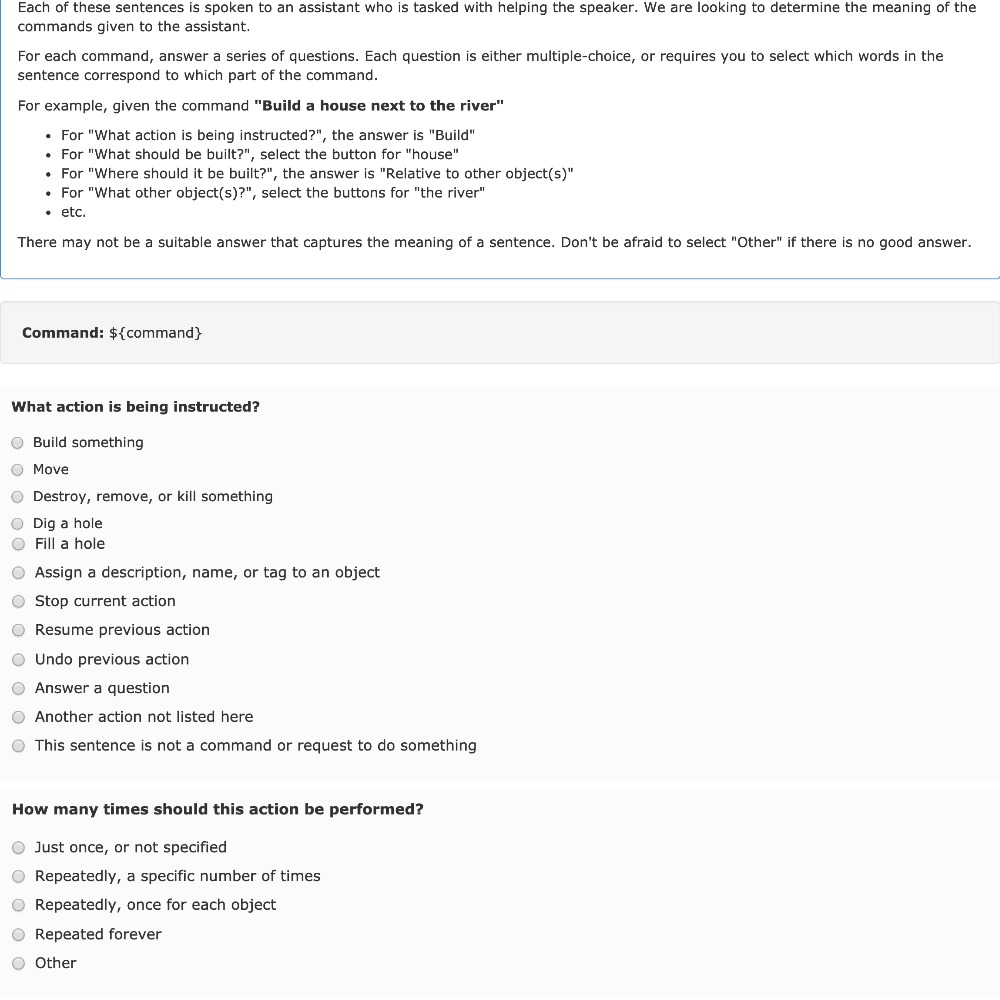}
	\caption{The task instructions shown to crowd-sourced workers for the annotation tool\label{fig:annotation_task}}
\end{figure}

The instructions shown to workers are shown in \ref{fig:annotation_task}.

%% file: confusion.tex
\clearpage

\section{Confusion Matrices}
\label{sec:conf_mat}

To compute confusion matrices for the internal node predictions, we look at the gold labels, and add:
\begin{itemize}
\item 1 to the gold label count when it is present in the predictions
\item $\frac{1}{\# predictions}$ for each predicted internal node that does not match a gold label node when the gold label is not present.
\end{itemize}
% Confusion matrices for internal nodes for the independent prediction and SentenceRec models respectively are presented in Figures~\ref{fig:conf_int_ind} and \ref{fig:conf_int_senrec} respectively.

To compute confusion matrices for the categorical node predictions, we look at the gold labels, and add:
\begin{itemize}
\item 1 to the predicted class count when the node is present in the predictions (whether it is the gold class or not)
\item 1 to the NO-PARENT result when the node's parent is absent in the predicted set
\item 1 to the ABSENT result when the node's parent is present in the predicted set but the categorical node is absent
\end{itemize}

To compute confusion matrices for the categorical node predictions, we look at the gold labels, and add:
\begin{itemize}
\item 1 to the MATCH-SPAN result count when the span node is present in the predictions with the right span
\item 1 to the MIS-SPAN result count when the span node is present in the predictions with the wrong span
\item 1 to the NO-PARENT result when the node's parent is absent in the predicted set
\item 1 to the ABSENT result when the node's parent is present in the predicted set but the categorical node is absent
\end{itemize}

\begin{figure}[t]
    \centering
    \begin{verbatim}
{'action': [('total', 817.0), ('action', 1.0)],
 'action_location': [('total', 73.0),
                     ('action_location', 0.8219),
                     ('ar_location', 0.1438),
                     ('schematic', 0.0137),
                     ('action_ref_object', 0.0137),
                     ('arl_ref_object', 0.0068)],
 'action_ref_object': [('total', 125.0),
                       ('action_ref_object', 0.64),
                       ('schematic', 0.16),
                       ('action_location', 0.0787),
                       ('al_ref_object', 0.0707),
                       ('stop_condition', 0.0347),
                       ('s_repeat', 0.012),
                       ('alr_repeat', 0.004)],
 'al_ref_object': [('total', 16.0),
                   ('al_ref_object', 0.875),
                   ('ar_location', 0.0938),
                   ('arl_ref_object', 0.0312)],
 'ar_repeat': [('total', 5.0),
               ('ar_repeat', 0.8),
               ('schematic', 0.1),
               ('s_repeat', 0.1)],
 's_repeat': [('total', 7.0),
              ('s_repeat', 0.8571),
              ('action_repeat', 0.0714),
              ('action_ref_object', 0.0714)],
 'schematic': [('total', 279.0),
               ('schematic', 0.9534),
               ('action_ref_object', 0.0376),
               ('ar_repeat', 0.0018),
               ('action_location', 0.0018),
               ('al_ref_object', 0.0018),
               ('ar_location', 0.0018),
               ('action_repeat', 0.0018)]}
    \end{verbatim}
    \caption{Confusion matrix for the internal node predictions by the SentenceRec model.}
    \label{fig:conf_int_senrec}
\end{figure}

\begin{figure}[t]
    \centering
    \begin{verbatim}
{'action:action_type': {'Answer': [('total', 52.0),
                                   ('Noop', 0.6923),
                                   ('Answer', 0.2308),
                                   ('Build', 0.0192),
                                   ('Dig', 0.0192),
                                   ('Destroy', 0.0192),
                                   ('Move', 0.0192)],
                        'Build': [('total', 358.0),
                                  ('Build', 0.8492),
                                  ('Noop', 0.0782),
                                  ('Spawn', 0.0559),
                                  ('Dig', 0.0168)],
                        'Destroy': [('total', 57.0),
                                    ('Destroy', 0.8246),
                                    ('Noop', 0.0702),
                                    ('Spawn', 0.0526),
                                    ('Move', 0.0175),
                                    ('Dig', 0.0175),
                                    ('Build', 0.0175)],
                        'Dig': [('total', 37.0),
                                ('Dig', 0.973),
                                ('Resume', 0.027)],
                        'Fill': [('total', 7.0),
                                 ('Fill', 0.5714),
                                 ('Build', 0.2857),
                                 ('Dig', 0.1429)],
                        'FreeBuild': [('total', 7.0),
                                      ('Build', 0.7143),
                                      ('Resume', 0.1429),
                                      ('Noop', 0.1429)],
                        'Move': [('total', 36.0),
                                 ('Move', 0.8611),
                                 ('Noop', 0.1389)],
                        'Noop': [('total', 113.0),
                                 ('Noop', 0.8053),
                                 ('Build', 0.0973),
                                 ('Move', 0.0265),
                                 ('Dig', 0.0265),
                                 ('Spawn', 0.0177),
                                 ('Answer', 0.0177),
                                 ('Fill', 0.0088)],
                        'OtherAction': [('total', 112.0),
                                        ('Noop', 0.5446),
                                        ('Move', 0.1786),
                                        ('Spawn', 0.0804),
                                        ('Dig', 0.0625),
                                        ('Destroy', 0.0536),
                                        ('Build', 0.0268),
                                        ('Fill', 0.0179),
                                        ('Undo', 0.0089),
                                        ('FreeBuild', 0.0089),
                                        ('Answer', 0.0089),
                                        ('Stop', 0.0089)],
    \end{verbatim}
    \caption{Confusion matrix for the categorical node predictions by the SentenceRec model part 1.}
    \label{fig:conf_cat_senrec_1}
\end{figure}

\begin{figure}[t]
    \centering
    \begin{verbatim}
                        'Resume': [('total', 3.0), ('Resume', 1.0)],
                        'Spawn': [('total', 16.0),
                                  ('Spawn', 0.875),
                                  ('Build', 0.125)],
                        'Stop': [('total', 10.0),
                                 ('Stop', 0.8),
                                 ('Move', 0.1),
                                 ('Destroy', 0.1)],
                        'Tag': [('total', 7.0),
                                ('Answer', 0.4286),
                                ('Build', 0.2857),
                                ('Noop', 0.1429),
                                ('Tag', 0.1429)],
                        'Undo': [('total', 2.0), ('Undo', 1.0)]},
 'action_location:location_type': {'AgentPos': [('NO-PARENT', 1.0),
                                                ('total', 1.0)],
                                   'BlockObject': [('total', 31.0),
                                                   ('NO-PARENT', 0.2903),
                                                   ('Mob', 0.2581),
                                                   ('AgentPos', 0.2581),
                                                   ('BlockObject', 0.1935)],
                                   'Other': [('total', 4.0),
                                             ('AgentPos', 0.5),
                                             ('SpeakerPos', 0.25),
                                             ('NO-PARENT', 0.25)],
                                   'SpeakerLook': [('total', 19.0),
                                                   ('NO-PARENT', 0.5789),
                                                   ('SpeakerLook', 0.4211)],
                                   'SpeakerPos': [('total', 37.0),
                                                  ('SpeakerPos', 0.7297),
                                                  ('NO-PARENT', 0.2703)]},
 'action_location:relative_direction': {'AWAY': [('total', 2.0), ('AWAY', 1.0)],
                                        'FRONT': [('total', 3.0),
                                                  ('NO-PARENT', 0.6667),
                                                  ('FRONT', 0.3333)],
                                        'LEFT': [('total', 2.0), ('LEFT', 1.0)],
                                        'UP': [('total', 2.0), ('UP', 1.0)]},
 'ar_repeat:repeat_key': {'FOR': [('total', 5.0),
                                  ('FOR', 0.8),
                                  ('NO-PARENT', 0.2)]},
 's_repeat:repeat_key': {'FOR': [('total', 7.0),
                                 ('FOR', 0.8571),
                                 ('NO-PARENT', 0.1429)]}}

    \end{verbatim}
    \caption{Confusion matrix for the categorical node predictions by the SentenceRec model part 2.}
    \label{fig:conf_cat_senrec_2}
\end{figure}

\begin{figure}[t]
    \centering
    \begin{verbatim}
{'action:has_depth_': [('total', 6.0),
                       ('ABSENT', 0.6667),
                       ('MIS-SPAN', 0.3333)],
 'action:has_size_': [('total', 9.0),
                      ('MATCH-SPAN', 0.6667),
                      ('ABSENT', 0.3333)],
 'action:has_width_': [('total', 2.0), ('MIS-SPAN', 1.0)],
 'action:tag': [('total', 7.0), ('ABSENT', 0.8571), ('MATCH-SPAN', 0.1429)],
 'action_ref_object:has_block_type_': [('total', 9.0), ('NO-PARENT', 1.0)],
 'action_ref_object:has_colour_': [('total', 2.0),
                                   ('MATCH-SPAN', 0.5),
                                   ('NO-PARENT', 0.5)],
 'action_ref_object:has_name_': [('total', 192.0),
                                 ('NO-PARENT', 0.5833),
                                 ('MATCH-SPAN', 0.3385),
                                 ('ABSENT', 0.0781)],
 'action_ref_object:has_size_': [('ABSENT', 1.0), ('total', 1.0)],
 'al_ref_object:has_name_': [('total', 29.0),
                             ('NO-PARENT', 0.5172),
                             ('MATCH-SPAN', 0.3448),
                             ('MIS-SPAN', 0.1379)],
 'ar_repeat:repeat_count': [('total', 5.0),
                            ('MATCH-SPAN', 0.8),
                            ('NO-PARENT', 0.2)],
 's_repeat:repeat_count': [('total', 7.0),
                           ('MATCH-SPAN', 0.8571),
                           ('NO-PARENT', 0.1429)],
 'schematic:has_block_type_': [('total', 37.0),
                               ('MATCH-SPAN', 0.7838),
                               ('ABSENT', 0.1351),
                               ('MIS-SPAN', 0.0541),
                               ('NO-PARENT', 0.027)],
 'schematic:has_colour_': [('total', 2.0), ('MATCH-SPAN', 1.0)],
 'schematic:has_height_': [('total', 7.0),
                           ('ABSENT', 0.5714),
                           ('MIS-SPAN', 0.2857),
                           ('NO-PARENT', 0.1429)],
 'schematic:has_name_': [('total', 336.0),
                         ('MATCH-SPAN', 0.7262),
                         ('NO-PARENT', 0.2083),
                         ('MIS-SPAN', 0.0625),
                         ('ABSENT', 0.003)],
 'schematic:has_size_': [('total', 21.0),
                         ('MATCH-SPAN', 0.5714),
                         ('NO-PARENT', 0.1905),
                         ('ABSENT', 0.1905),
                         ('MIS-SPAN', 0.0476)],
 'schematic:has_width_': [('total', 7.0),
                          ('ABSENT', 0.8571),
                          ('NO-PARENT', 0.1429)]}

    \end{verbatim}
    \caption{Confusion matrix for the span node predictions by the SentenceRec model.}
    \label{fig:conf_int_senrec}
\end{figure}